\documentclass{article}

\usepackage{arxiv}

\usepackage[utf8]{inputenc} 
\usepackage[T1]{fontenc}    
\usepackage[bookmarks=false]{hyperref}       
\usepackage{url}            
\usepackage{booktabs}       
\usepackage{amsfonts}       
\usepackage{nicefrac}       
\usepackage{microtype}      
\usepackage{lipsum}
\usepackage{graphicx}
\usepackage{longtable}
\usepackage{glossaries}
\usepackage{enumerate}
\usepackage[export]{adjustbox}
\usepackage{amssymb}
\usepackage{pifont}
\usepackage{pbox}

\makeglossaries

\newacronym{AL}{AL}{Average Linkage}
\newacronym{ARIMA}{ARIMA}{Auto-regressive Integrated Moving Average}
\newacronym{ARR}{ARR}{Adjusted Ratio of Ratios}
\newacronym{AverageNodes}{Average Nodes}{Average Nodes Learner}
\newacronym{Base-learning}{BLL}{Base-level Learning}
\newacronym{BinaryFeatures}{b}{Number of Binary Features}
\newacronym{C4.5}{C4.5}{C4.5 Decision Tree algorithm}
\newacronym{C5.0boost}{C5.0 boost}{C5.0 Adaptive Boosting}
\newacronym{C5.0rules}{C5.0 rules}{C5.0 Rule Induction}
\newacronym{C5.0trees}{C5.0 tree}{C5.0 Decision Tree}
\newacronym{CANCOR}{CANCOR}{Canonical Correlation}
\newacronym{CART}{CART}{Classification and Regression Trees}
\newacronym{CASTLE}{CASTLE}{Causal Structure for Inductive Learning}
\newacronym{CBR}{CBR}{Case-based Reasoning}
\newacronym{CL}{CL}{Complete Linkage}
\newacronym{Classes}{k}{Number of Classes}
\newacronym{CN2}{CN2}{CN2 Induction Algorithm}
\newacronym{CORR}{CORR}{Mean Absolute Correlation Coefficient}
\newacronym{CoV}{CoV}{Coefficient of Variation}
\newacronym{CV}{CV}{Cross-Validation}
\newacronym{DBS}{DBS}{DB-Scan}
\newacronym{DecisionNodes}{Decision Nodes}{Decision Nodes Learner}
\newacronym{DCT}{DCT}{Dataset Characterization Tool}
\newacronym{DSIT}{DSIT}{Descriptive, Statistical and Information-Theoretic}
\newacronym{DT}{DT}{Decision Trees}
\newacronym{DiscFunc}{DiscFunc}{Number of Discriminant Functions}
\newacronym{DMA}{DMA}{Data Mining Advisor}
\newacronym{DW}{DW}{Durbin-Watson statistic of regression residual}
\newacronym{e-LICO}{e-LICO}{e-Laboratory for Interdisciplinary Collaborative Research}
\newacronym{e-NN}{e-NN}{Elite-Nearest Neighbour}
\newacronym{ES}{ES}{Exponential Smoothing}
\newacronym{EWS}{EWS}{Early Warning System}
\newacronym{EoD}{EoD}{Examples of Datasets}
\newacronym{FFT}{FFT}{Fast Fourier Transform}
\newacronym{Features}{p}{Number of Features}
\newacronym{FDA}{FDA}{Fisher Discriminatory Analysis}
\newacronym{FF}{FF}{Farthest First}
\newacronym{FLD}{FLD}{Fisher's Linear Discriminant}
\newacronym{FRACT}{FRACT}{Relative proportion of largest Eigenvalue}
\newacronym{HC}{HC}{Entropy of Classes}
\newacronym{HCX}{HCX}{Joint Entropy of Classes}
\newacronym{HX}{HX}{Average Entropy}
\newacronym{IBL}{IBL}{Instance-based Learning}
\newacronym{ICA}{ICA}{Independent Component Analysis}
\newacronym{ID3}{ID3}{Iterative Dichotomiser 3}
\newacronym{IDA}{IDA}{Intelligent Discovery Assistant}
\newacronym{INDCART}{INDCART}{Inductive CART}
\newacronym{Instances}{N}{Total Instances}
\newacronym{k-M}{k-M}{k-Means}
\newacronym{k-NN}{k-NN}{k-Nearest Neighbour}
\newacronym{KD}{KD}{Knowledge Discovery}
\newacronym{KURT}{KURT}{Kurtosis}
\newacronym{LazyDT}{LazyDT}{Lazy Decision Trees}
\newacronym{Ltree}{Ltree}{Linear Discriminant Trees}
\newacronym{LDA}{LDA}{Linear Discriminant Analysis}
\newacronym{LVQ}{LVQ}{Learning Vector Quantization}
\newacronym{LWPR}{LWPR}{Locally Weighted Projection Regression}
\newacronym{M}{M}{Mixture Models}
\newacronym{MA}{MA}{Moving Average}
\newacronym{MAE}{MAE}{Mean Absolute Error}
\newacronym{MCX}{MCX}{Average Mutual Information between Class and Nominal Features}
\newacronym{MDS}{MDS}{Multi-dimensional Scaling}
\newacronym{METAL}{METAL}{Meta-Learning Assistant}
\newacronym{METALA}{METALA}{Meta-learning Architecture}
\newacronym{Meta-example}{ME}{Meta-example}
\newacronym{Meta-features}{MF}{Meta-feature}
\newacronym{Meta-knowledge}{MK}{Meta-knowledge}
\newacronym{Machine Learning}{ML}{Machine Learning}
\newacronym{MARS}{MARS}{Multivariate Adaptive Regression Splines}
\newacronym{Meta-learning}{MLL}{Meta-level Learning}
\newacronym{MLP}{MLP}{Multi-layer Perceptron}
\newacronym{MLR}{MLR}{Multiple Linear Regression}
\newacronym{MLT}{MLT}{Machine Learning Toolbox}
\newacronym{MSE}{MSE}{Mean Squared Error}
\newacronym{NB}{NB}{Naive Bayes classifier}
\newacronym{NBT}{NBT}{Naive Bayes/Decision-Tree}
\newacronym{NN}{NN}{Neural Network}
\newacronym{NominalFeatures}{s}{Number of Nominal features}
\newacronym{NoiseRaio}{NoiseRaio}{Noise to Signal Ratio}
\newacronym{NumericFeatures}{n}{Number of Numeric features}
\newacronym{OC1}{OC1}{Oblique Classifier-1}
\newacronym{OneR}{OneR}{One Rule Learner}
\newacronym{PaREn}{PaREn}{Pattern Recognition Engineering}
\newacronym{PEBLS}{PEBLS}{Parallel Exemplar-Based Learning System}
\newacronym{PCA}{PCA}{Principal Component Analysis}
\newacronym{PHT}{PHT}{Page-Hinkley Test}
\newacronym{PPR}{PPR}{Projection Pursuit Regression}
\newacronym{RW}{RW}{Random Walk}
\newacronym{RandomlyNodes}{Randomly Chosen Nodes}{Randomly Chosen Nodes Learner}
\newacronym{RapidAnalytics}{RapidAnalytics}{open-source data-mining and predictive analysis solution}
\newacronym{RBF}{RBF}{Radial-basis Function}
\newacronym{RF}{RF}{Random Forests}
\newacronym{RL}{RL}{Reinforcement Learning}
\newacronym{Ripper}{Ripper}{Rule Learner}
\newacronym{RMSE}{RMSE}{Root Mean Squared Error}
\newacronym{SDRatio}{S/D Ratio}{Homogeneity of Covariances}
\newacronym{SKEW}{SKEW}{Skewness}
\newacronym{SL}{SL}{Single Linkage}
\newacronym{SMAPE}{SMAPE}{Symmetric Mean Absolute Percentage Error}
\newacronym{SMART}{SMART}{Smooth Multiple Additive Regression Technique}
\newacronym{SMOTE}{SMART}{Synthetic Minority Over-sampling TEchnique}
\newacronym{SNN}{SNN}{Shared Nearest Neighbours}
\newacronym{SP}{SP}{Spectral Clustering}
\newacronym{Spearman}{SRCC}{Spearman's Rank Correlation Coefficient}
\newacronym{STABB}{STABB}{Shift To A Better Bias}
\newacronym{StatLog}{StatLog}{Statistical and Logical learning}
\newacronym{StdDev}{StdDev}{Standard Deviation}
\newacronym{SVM}{SVM}{Support Vector Machines}
\newacronym{SVR}{SVR}{Support Vector Regression}
\newacronym{Time-series}{TS}{Time-series}
\newacronym{TestInstances}{t}{Number of Test instances}
\newacronym{TrainingInstances}{r}{Number of Training instances}
\newacronym{UCI}{UCI}{UCI Machine Learning Repository}
\newacronym{QPC}{QPC}{Quality of Projected Clusters}
\newacronym{Quadra}{Quadra}{Quadratic Classifier}
\newacronym{VBMS}{VBMS}{Variable-bias Management System}
\newacronym{Wlambda}{Wlambda}{Wilks'lambda Distribution}
\newacronym{WorstNodes}{Worst Nodes}{Worst Nodes Learner}
\newacronym{XM}{XM}{X-Means}

\newcommand{\tick}{\ding{52}}

\title{A Review of Meta-level Learning in the Context of Multi-component, Multi-level Evolving Prediction Systems}

\author{
 Abbas Raza Ali \\
  Faculty of Science and Technology\\
  Bournemouth University\\
  Poole BH12 5BB \\
  United Kingdom \\
  \texttt{aali@bournemouth.ac.uk} \\
  \And
 Marcin Budka \\
  Faculty of Science and Technology\\
  Bournemouth University\\
  Poole BH12 5BB \\
  United Kingdom \\
  \texttt{mbudka@bournemouth.ac.uk} \\
   \And
 Bogdan Gabrys \\
  Advanced Analytics Institute\\
  University Technology Sydney\\
  Ultimo NSW 2007\\
  Australia\\
  \texttt{bogdan.gabrys@uts.edu.au} \\
}

\begin{document}
\maketitle

\begin{abstract}
The exponential growth of volume, variety and velocity of data is raising the need for investigations of automated or semi-automated ways to extract useful patterns from the data. It requires deep expert knowledge and extensive computational resources to find the most appropriate mapping of learning methods for a given problem. It becomes a challenge in the presence of numerous configurations of learning algorithms on massive amounts of data. So there is a need for an intelligent recommendation engine that can advise what is the best learning algorithm for a dataset. The techniques that are commonly used by experts are based on a trial and error approach evaluating and comparing a number of possible solutions against each other, using their prior experience on a specific domain, etc. The trial and error approach combined with the expert's prior knowledge, though computationally and time expensive, have been often shown to work for stationary problems where the processing is usually performed off-line. However, this approach would not normally be feasible to apply to non-stationary problems where streams of data are continuously arriving. Furthermore, in a non-stationary environment, the manual analysis of data and testing of various methods whenever there is a change in the underlying data distribution would be very difficult or simply infeasible. In that scenario and within an on-line predictive system, there are several tasks where Meta-learning can be used to effectively facilitate best recommendations including 1) pre-processing steps, 2) learning algorithms or their combination, 3) adaptivity mechanisms and their parameters, 4) recurring concept extraction, and 5) concept drift detection. However, while conceptually very attractive and promising, the Meta-learning leads to several challenges with the appropriate representation of the problem at a meta-level being one of the key ones. 

The goal of this review and our research is, therefore, to investigate Meta-learning in general and the associated challenges in the context of automating the building, deployment and adaptation of multi-level and multi-component predictive system that evolves over time.
\end{abstract}

\keywords{Adaptive Mechanisms \and Domain Adaption \and Meta-features \and Meta-knowledge \and Meta-learning \and Transfer Learning}

\section{Introduction}
One of the major challenges in \gls{Machine Learning} is to predict when one algorithm is more appropriate than another to solve a learning problem~\cite{Prudencio2011}. Traditionally, estimating the performance of algorithms has involved an intensive trial-and-error process which often demands massive execution time and memory together with the advice of experts that are not always easy to acquire~\cite{Giraud-Carrier2004}. \gls{Meta-learning} has been identified as a potential solution to this problem~\cite{lemke2013metalearning}. It uses examples from various domains to produce a \gls{Machine Learning} model, known as a Meta-learner, which is responsible for associating the characteristics of a problem with the most appropriate candidate algorithms found to have worked best on previously solved similar problems. The knowledge which is used by a Meta-learner is acquired from previously solved problems, where each problem is characterized by several features, known as \glspl{Meta-features}. \glspl{Meta-features} are combined with performance measures of learning algorithms to build a \gls{Meta-knowledge} database. Learning at the base-level gathers experience within a specific problem, while \gls{Meta-learning} is concerned with accumulating experience over several learning problems~\cite{Giraud-Carrier2008}.

\gls{Meta-learning} started to appear in the \gls{Machine Learning} literature in the 1980s and was referred to by different names like dynamic bias selection~\cite{Rendell1987}, algorithm recommender~\cite{Brazdil2008}, etc. Sometimes \gls{Meta-learning} is also used with a reference to ensemble methods~\cite{Duch2011} which can cause some confusion. So, in order to get a comprehensive view of exactly what \gls{Meta-learning} is, a number of definitions have been proposed in various studies. \cite{Vilalta2002} and~\cite{Vanschoren2011} define \gls{Meta-learning} as the understanding of how learning itself can become flexible according to the domain or task and how it tends to adapt its behaviour to perform better. \cite{Giraud-Carrier2008} describes it as the understanding of the interaction between the mechanism of learning and concrete context in which that mechanism is applicable. \cite{Brazdil2008} view on \gls{Meta-learning} is that it is the study of methods that exploit Meta-knowledge to obtain efficient models and solutions by adapting the learning algorithms, while \gls{Meta-knowledge} is a combination of characteristics and performance measures of \gls{EoD}. To some extent, this definition is followed in this research as well.

Extracting \glspl{Meta-features} from a dataset plays a vital role in the \gls{Meta-learning} task. Several \gls{Meta-features} generation approaches are available to extract a variety of information from previously solved problems. The most commonly used approaches are descriptive (or simple), statistical, information-theoretic, landmarking and model-based. The \gls{DSIT} features are easy to extract from the dataset as compared to the other approaches. Most of them have been proposed in the same period of time and are often used together in most of the studies. These approaches are used to assess the similarity of a new dataset to previously analysed datasets~\cite{Bensusan2000a}. Landmarking is the most recent approach that tries to relate the performance of candidate algorithms to the performance obtained by simpler and computationally more efficient learners~\cite{Pfahringer2000}. The Model-based approach attempts to capture the characteristics of a problem from the structural shape and size of a model induced from the dataset~\cite{Peng2002}. The decision tree models are mostly used in this approach, where properties are extracted from the tree, such as tree depth, shape, nodes per feature, etc. \cite{Giraud-Carrier2008}.

The \gls{Meta-features} extraction approaches listed above are used in several implementations of decision-support systems for algorithm selection. One of the initial studies to address the practice of \gls{Meta-learning} was \gls{MLT} project by~\cite{Graner1994}. The project was a kind of expert system for algorithm selection which gathered user inputs through a set of questions about the data, the domain and user preferences. Although its knowledge-base was built through expert-driven knowledge engineering rather than via \gls{Meta-learning}, it still stood out as the first automatic tool that systematically relates application domain and dataset characteristics. In the same period,~\cite{King1995} contributed with statistical and information-theoretic measures based approach for classification tasks, known as \gls{StatLog}. A large number of \glspl{Meta-features} were used in \gls{StatLog} together with a broad class of candidate models for the algorithm selection task. The project produced a thorough empirical analysis of various classifiers and learning models using different performance measures. \gls{StatLog} was followed by various other implementations with some refinement in \gls{Meta-features} set, input datasets, \gls{Base-learning} and \gls{Meta-learning} algorithms. An EU funded research project \gls{METAL} had a key objective to facilitate a selection of the best-suited classification algorithm for a data-mining task~\cite{Berrer2000}. \gls{METAL} introduced new relevant \glspl{Meta-features} and ranked various classifiers using statistical and information-theoretic approaches. A ranking mechanism was also proposed by exploiting the ratio of accuracy and training time. An agent-based architecture for distributed Data Mining, \gls{METALA}, was proposed in~\cite{Botia2001}. Its aim was the automatic selection of an algorithm that performs best from a pool of available algorithms by automatically carrying out experiments with each learner and task to induce a Meta-model for algorithm selection. The \gls{IDA} provided a knowledge discovery ontology that defined the existing techniques and their properties~\cite{Bernstein2001}. \gls{IDA} used three algorithmic steps of the knowledge discovery process, which included: 1) pre-processing, 2) data modelling, and 3) post-processing. It generated all valid processes and then a heuristic ranker could be applied to compute user-specified goals which were initially gathered as input. Later,~\cite{Bernstein2005} research focused on extending~\cite{Bernstein2001} approach by leveraging the interaction between ontology to extract deep knowledge and case-based reasoning for \gls{Meta-learning}. One of the recent contributions to \gls{Meta-learning} practice was made by~\cite{Mierswa2006} under \gls{PaREn} project. A Landmarking operator was one of the outcomes of this project which was later embedded in RapidMiner. These systems are described in more detail in Section~\ref{sec:metaLearning}.

While there has been a lot of interest in \gls{Meta-learning} approaches and significant progress has been made, there are a number of outstanding issues that will be explained and some of which will be addressed. The main challenge of this work is research on \gls{Meta-learning} strategies and approaches in the context of adaptive multi-level, multi-component predictive systems. This problem leads to several research challenges and questions which are discussed in detail in Section~\ref{sec:challenges}.

\subsection{The review context and the INFER project summary}
The research described in this report is closely related to and was conducted within the framework of the recently completed INFER\footnote{\url{http://infer.eu/}} project.
INFER stands for Computational Intelligence Platform for Evolving and Robust Predictive Systems and was a project funded by the European Commission within the Marie Curie Industry-Academia Partnerships \& Pathways (IAPP) programme with a runtime from July 2010 until June 2014.

INFER project's research programme and partnership focused on pervasively adaptive software systems for the development of an open, modular software platform for predictive modelling applicable in different industries and a next generation of adaptive soft sensors for on-line prediction, monitoring and control in the process industry.

The main project goals were achieved by pursuing the following objectives within three overlapping research and partnership programme areas:

\begin{enumerate}[1.]
    \item Computational Intelligence – Objective 1: Research and development of advanced mechanisms for adaptation, increased robustness and complexity management of highly flexible, multi-component, multi-level evolving predictive systems.
    
    \item Software Engineering – Objective 2: Development of professionally coded INFER software platform for robust predictive systems building and intelligent data analysis.
    
    \item Process Industry/Control Engineering – Objective 3: Development of self-adapting and monitoring soft sensors for the process industry.
\end{enumerate}

When the project was starting in 2010, there were several freely accessible general-purpose data mining and intelligent data analysis software packages and libraries on the market which could be used to develop predictive models, but one of their main drawbacks was that advanced knowledge of how to select and configure available algorithms was required. A number of commercial data mining/predictive modelling software packages were also available. These tools attempted to automate some steps of the modelling process (e.g. data pre-processing, handling of missing values or even model complexity selection) thus reducing the required expertise of the user. Most of them were however either front-ends for a single data mining/machine learning technique or they were specialised tools designed specifically for use by a single industry. All these tools had one thing in common – generated models were static and the lack of full adaptability implied the need for their periodic manual tuning or redesign.

The main innovation of the INFER project was therefore the creation and investigation of a novel type of environment in which the ‘fittest’ predictive model for whatever purpose would emerge – either autonomously or by user high-level goal-related assistance and feedback. In this environment, the development of predictive systems would be supported by a variety of automation mechanisms, which would take away as much of the model development burden from the user as possible. Once applied, the predictive system should be able to exploit any available feedback for its performance monitoring and adaptation.

There were (and still are) a lot of fundamental research questions related to the automation of data-driven predictive models building, ensuring their robust behaviour and development of integrated adaptive/learning algorithms and approaches working on different time scales from real-time adaptation to life long learning and optimisation. All of these questions provided the main thrust of advanced research conducted in the project and resulted in contributions to a large number (over 70) of high impact publications in top journals and international conferences. While all of the papers can be accessed via the project website (http://www.infer.eu) some of the key ones related to this review are listed below for easy access and reference.
We split the publications using a set of distinct areas of interest and investigation and combine both the the older ones which led to the conception of the project in the first place and some which resulted from running the project. These are: i. complex adaptive systems and architectures (\cite{gabrys2005smart,ruta2007neural,kadlec2009architecture,zliobaite2012next}); ii. classifier and predictor ensembles (\cite{ruta2002theoretical, gabrys2002combining, gabrys2004learning, ruta2005classifier, gabrys2006genetic, ruta2007neural, riedel2007dynamic, budka2010ridge, eastwood2012generalised}); iii. multi-level and multi-component predictors (\cite{ruta2002theoretical,riedel2005evolving,riedel2005hierarchical,riedel2007combination,riedel2009pooling, tsakonas2012gradient, lemke2013evolving, tsakonas2013fuzzy});
iv. meta-learning (\cite{LemkeJun2010,LemkeJul2010,lemke2013metalearning}, v. learning and adaptation in changing environments (\cite{sahel2007adaptive,kadlec2011review, tsakonas2011evolving, bakirov2013investigation, Gama2014,zliobaite2014adaptive}); vi. representative data sampling and predictive model evaluation (\cite{budka2010correntropy,budka2011accuracy,budka2013density}); vii. adaptive soft sensors (\cite{kadlec2008adaptive, kadlec2008gating, kadlec2008soft, kadlec2009data, kadlec2009evolving, kadlec2009soft, kadlec2010adaptive, kadlec2011local, kadlec2011review, budka2014sensor}) and viii. other application areas (\cite{lemke2008we,lemke2009dynamic, stahl2013overview, salvador2014online}). 

A variety of application areas and contexts have been used to illustrate the performance of developed approaches and/or to understand the mechanisms governing their behaviour. One of the key applications considered and tackled was that of adaptive soft sensors needed in the process industry.

The INFER software platform, developed with the creation of highly flexible, multi-component, multi-level evolving predictive systems in mind, supports parallel training, validation and execution of multiple predictive models, with each of them potentially being in a different state. Moreover, various optimization tasks can also be run in the background, taking advantage of idle computational resources. The predictive models running within the INFER platform are inherently adaptive. This means that they constantly evolve towards more optimal solutions as new data arrives. The importance of this feature stems from the fact, that real data is seldom stationary – it often undergoes various changes, which affect the relationships between inputs and outputs, rendering fixed predictive models unusable. A distinguishing feature of the INFER software is an intelligent automation of the predictive model building process, allowing non-experts to create well-performing and robust predictive systems with minimal effort. At the same time, the system offers full flexibility for the expert users in terms of the choice, parameterisation and operation of the predictive methods as well as efficient integration of domain knowledge. While there is still a substantial development effort required before a viable commercial software product could be delivered the strong foundations have been created and it is our intention to build on them in the future.

More information on the INFER\footnote{\url{http://infer.eu/}} project and its outcomes can be found following the link in the footnote. 

The rest of the report is structured as follows. The next chapter covers the existing research in \gls{Meta-learning} area, including some important components of an \gls{Meta-learning} system. Those components include 1. the sources of existing and ways of automatic generation of datasets, 2. Meta-feature generation and selection using various approaches, and 3. base-level learning algorithms performance measures, such as accuracy, execution time, etc. This is followed by sections discussing existing Meta-learning systems in the context of their applicability to the supervised and unsupervised algorithms. The last section of Section~\ref{sec:existing_research} illustrates the adaptive mechanism aspect in detail. Based on the conclusions and recommendations extracted from the literature review, Section~\ref{sec:challenges} describes the research challenges of this work in the context of multi-component and multi-level adaptive systems. 
And finally, the summary is provided in Section~\ref{sec:summary}.

\section{Existing Research} \label{sec:existing_research}
\begin{figure}[h!] 
    \centering
	\includegraphics[scale = 0.85, center]{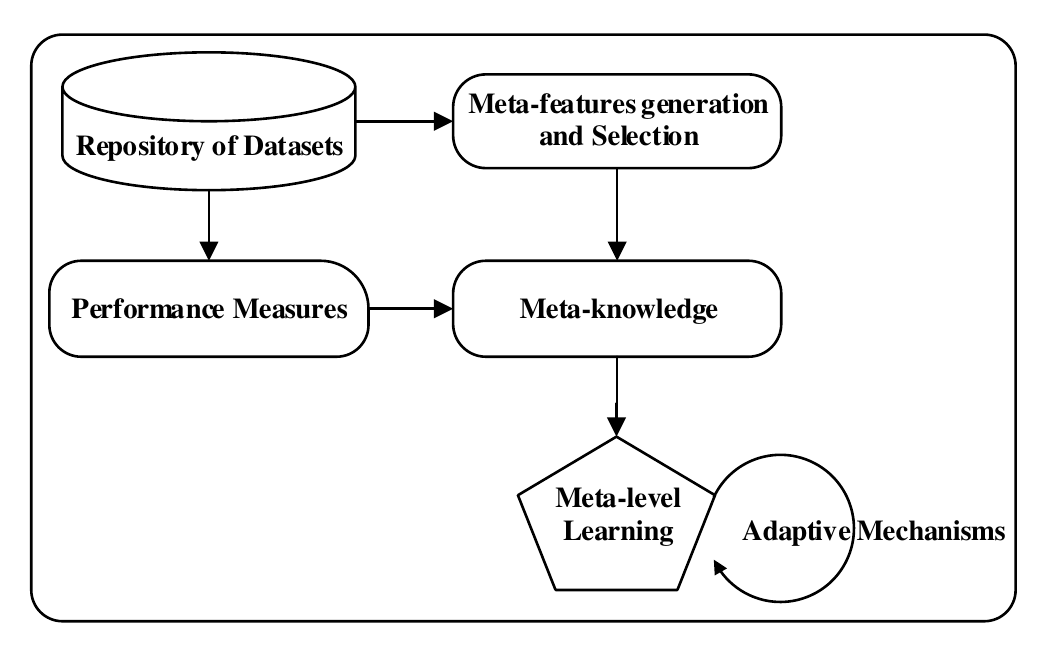} 
	\caption{Scope of existing research review}
	\label{fig:FigExistingResearch}
\end{figure}

A lot of research has been conducted on automating \gls{Machine Learning} algorithm selection in the last three decades. The focus of many of those studies is on various components of \gls{Meta-learning}. Because of our particular interest in MLL, the scope of this literature review is confined to areas that are directly related to the MLL research. The high-level overview of the components which are discussed in this chapter is shown in Figure~\ref{fig:FigExistingResearch}. The first section is discussing ways of gathering real-world datasets and techniques to create synthetic datasets which are known as \gls{EoD}. These \gls{EoD} are used to generate \glspl{Meta-features} and associated performance measures which are discussed in Sections 2.2 and 2.3 respectively. \gls{Meta-features} are combined with performance measures to build \gls{Meta-knowledge} dataset which becomes the input of \gls{Meta-learning}. The last section illustrates adaptive mechanisms in the context of \gls{Meta-learning} which are an important aspect of our research focus. 

\subsection{Repository of Datasets} \label{sec:datasets}
A repository of datasets representing various problems is one of the key components of the entire \gls{Meta-learning} system. As \cite{Vanschoren2011} states, there is no lack of experiments being done, but the datasets and information obtained often remain in the \emph{people's heads and labs}. This section explores the sources of real-world datasets that are used in the existing studies to build \gls{Meta-knowledge} database. However, real-world datasets are usually hard to obtain but artificially generated datasets would be a possible solution to this problem. In the following subsections, studies that are dealing with the real-world data, those which elaborate the techniques to generate artificial datasets, and published resources are discussed.

\subsubsection{Real-world Datasets}
The real-world datasets can be difficult to find and gather in the desired format. An effort has been made to extract useful sources of data from various studies. Table~\ref{table:ExamplesofDatasets} presents datasets that are used in different researches for \gls{Meta-learning} purpose. Most of them are gathered from \gls{UCI} \cite{Bache2013}. 

\begin{center}
{\small
\begin{longtable}{c|p{2.1cm}|p{7.3cm}|p{3.5cm}}
\caption{Real-world datasets used in various studies}
\label{table:ExamplesofDatasets} \\ 
\toprule
Research Work & Datasets & Sources & Dataset Filters \\
\midrule
\cite{King1995} & 12 & Satellite image, Hand written digits, Karhunen-Loeve digits, Vehicle silhouettes, Segment data, Credit risk, Belgian data, Shuttle control, Diabetes, Heart disease, German credit, Head injury \cite{KingUCI1995} & - \\
\cite{Lindner1999} & 80 & \gls{UCI} and DaimlerChrysler & - \\
\cite{Sohn1999} & 19 & Satellite image, Hand written digits, Karhunen-Loeve digits, Vehicle silhouettes, Segment data, Credit risk, Belgian data, Shuttle control, Diabetes, Heart disease, German credit, Head injury \cite{KingUCI1995} and 7 other datasets used in StatLog project & Three datasets of StatLog having cost information involved in misclassification \\
\cite{Berrer2000} & 58 & \gls{METAL} project datasets & 38 datasets with no missing values \\
\cite{Soares2001} & 45 & \gls{UCI} and DaimlerChrysler & Dataset with more than 1000 instances \\
\cite{Bernstein2001} & 15 & Balance Scale, Breast Cancer, Heart disease, Heart disease - compressed glyph visualization, German Credit Data, Diabetes, Vehicle silhouettes, Horse colic, Ionosphere, Vowel, Sonar, Anneal, Australian credit data, Sick, Segment data \cite{Bache2013} & - \\
\cite{Todorovski2002} & 65 & \gls{UCI} and \gls{METAL} project datasets & 38 datasets with no missing values \\
\cite{Brazdil2003} & 53 & \gls{UCI} and DaimlerChrysler & Datasets with more than 100 instances \\
\cite{Bernstein2005} & 23 & Balance Scale, Heart disease, Heart disease, Heart disease - compressed glyph visualization, German Credit Data, Diabetes, Vehicle silhouettes, Ionosphere, Vowel, Anneal, Australian credit data, Sick, Segment data, Robot Moves, DNA, Gene, Adult 10, Hypothyroid, Waveform, Page blocks, Optical digits, Insurance, Letter, Adult \cite{Bache2013} & - \\
\cite{Peng2002} & 47 & \gls{UCI} & - \\
\cite{Kopf2002} & 78 & \gls{UCI} & Dataset with less than 1066 instances and the number of attributes ranged from 4 to 69 \\
\cite{Prudencio2004} & I: 99 \gls{Time-series} \& II: 645 & I: Time-series Data Library\footnote{\url{http://datamarket.com/data/list/?q=provider:tsdl}} and II: M3 competition\footnote{\url{http://forecasters.org/resources/time-series-data/m3-competition}} & I: Stationary data and II: Yearly data \\
\cite{Prudencio2008} & 50 & WEKA project\footnote{Machine Learning Group at University of Waikato \url{http://www.cs.waikato.ac.nz/ml/weka}} & On average datasets contain 4,392 instances and 14 features \\

\cite{Wang2009} & 46 and 5 & Time Series Data-mining Archive\footnote{\url{http://www.cs.ucr.edu/~eamonn/time_series_data}} and Time Series Data Library\footnote{\url{http://datamarket.com/data/list/?q=provider:tsdl}} & - \\

\cite{kadlec2009architecture} & 3 & Thermal oxidiser, Industry drier, and Catalyst activation datasets of process industry & On-line prediction datasets \\

\cite{LemkeJun2010} & 2 consisting of 111 \gls{Time-series} & NN3\footnote{Neural Network forecasting competition}\addtocounter{footnote}{-1}\addtocounter{Hfootnote}{-1} - Monthly business with 52-126 observations and NN5\footnotemark - daily cash machine withdrawals with 735 observations in each series & NN5 including some missing values \\

\cite{Abdelmessih2010} & 90 & \gls{UCI} & Datasets with more than 100 instances \\

\cite{Duch2011} & 5 and 2 & Leukemia, Heart, Wisconsin, Spam, and Ionosphere are real-world datasets gathered from \gls{UCI} and two synthetic datasets parity and monks & - \\

\cite{Rossi2014} & 2 & Travel Time Prediction (TTP) consists of 24,975 instances and Electricity Demand Prediction (EDP) consists of 27,888 instances & - \\
\bottomrule
\end{longtable}}
\end{center}

\cite{Warden2011} wrote a concise handbook that covers the most useful sources of publicly available datasets. A lot of new sources of free and publicly available data that have emerged over the last few years are discussed. Apart from discussing data-sources, methods to get datasets in bulk from those sources are also discussed in detail. Table~\ref{table:SourcesofDatasets} presents most of the sources from the author's book.

\begin{center}
{\small
\begin{longtable}{p{2.7cm}|p{7.0cm}|p{1.7cm}|p{3.2cm}}
\caption{List of publicly available Data Repositories}
\label{table:SourcesofDatasets} \\
\toprule
Source & Description & Datasets & Industry \\
\midrule
AnalcatData & Datasets that are used by Jeffrey S. Simonoff in his book \emph{Analyzing Categorical Data}, published in July 2003 & 83 & Cross-industry \\
Amazon Web Services & A centralized repository of public datasets & - & Astronomy, Biology, Chemistry, Climate, Economics, Geographic and Mathematics \\
Bioassay data & Virtual screening of bioassay (active/inactive compounds) data by Amanda Schierz & 21 & Life Sciences \\
Canada Open Data & Canadian government and geospatial data & - & Government \& Geospatial \\
Datacatalogs & List of the most comprehensive open data catalogs & - & - \\
data.gov.uk & Data of UK central government departments, other public sector bodies and local authorities & 9616 & Government and Public Sector \\
data.london.gov.uk & Data of UK central government departments, other public sector bodies and local authorities & 563 & Government and Public sector \\
Data.gov/Education & Educational high-value datasets & 70,897 & Cross-industry \\
ELENA & Non-stationary streaming data of flight arrival and departure details for all the commercial flights within the USA & 13 features and 116 million instances & Aviation \\
KDD Cup & Annual Data Mining and Knowledge Discovery competition datasets & - & cross-industry \\
Open Data Census US Census Bureau & Assesses the state of open data around the world & - & Government and Public sector \\
OpenData from Socrata & Freely available datasets & 10,000 & Business, Education, Government, Social and Entertainment \\
Open Source Sports & Many sports databases, including Baseball, Football, Basketball and Hockey & - & Entertainment \\
\gls{UCI} & A collection of databases, domain theories, and data generators that are used by the \gls{Machine Learning} community for the empirical analysis of learning algorithms & 199 & Physical Sciences, Computer Science \& Engineering, Social Sciences, Business and Game \\
Yahoo Sandbox datasets & Language, graph, ratings, advertising and marketing, competition, computing systems and image datasets & - & Cross-industry \\
\bottomrule
\end{longtable}}
\end{center}

\subsubsection{Synthetic Datasets} 
\glspl{Meta-features} are used as predictors in an \gls{Meta-learning} system. Typically, many \glspl{Meta-features} are extracted from a dataset, thereby leading to a high-dimensional sparsely populated feature space which has always been a challenge for learning algorithms. Hence, to overcome this problem sufficient number of datasets are required which may not be possible only from the repositories of the real-world datasets as they can be hard to obtain. So, artificially generated datasets might help in solving this issue. \cite{Rendell1990} work on systematic artificial data generation is considered as one of the initial efforts in this regard. 

\cite{Bensusan2000b} used 320 artificially generated boolean datasets with 5 to 12 features in each one. These artificial datasets were benchmarked on 16 \gls{UCI} and DaimlerChrysler real-world datasets. Similarly \cite{Pfahringer2000} generated 222 datasets, each containing 20 numeric and nominal features having 1K to 10K instances classified between 2 to 5 classes. Additionally, 18 real-world \gls{UCI} problems were used to evaluate the proposed approach. 

\cite{Soares2009Apr} proposed a method to generate a large number of datasets by transforming the existing datasets, known as \emph{datasetoids}. An artificial dataset was generated against each symbolic attribute of a given dataset, obtained by switching the selected attribute with the target variable. This method was used on 64 datasets gathered from the \gls{UCI} repository and it generated a total of 983 \emph{datasetoids}. At the end potential anomalies related to the artificial datasets were also discussed as well as their proposed solutions were presented. Those identified anomalies were: 1) the new target variable having missing values, 2) the target variable being very skewed, and/or 3) the corresponding target variable being completely unrelated to the remaining features. One very simple solution proposed for these problems was to simply discard the \emph{datasetoids} which showed any of the above mentioned properties. This method produced promising results, therefore enabling the generation of new datasets that could solve the scarce datasets problems.

\cite{Wang2009} used both synthetic and real-world Time Series (\gls{Time-series}) from diverse domains for \gls{Meta-learning} based forecasting method selection study. The details of real-world datasets are given in Table~\ref{table:ExamplesofDatasets} while remaining synthetic datasets were generated using statistical simulation to facilitate the detailed analysis of forecasting association with data characteristics. A total of 264 artificial datasets were generated to exhibit a number of different characteristics including, for instance, perfect and strong trend, perfect seasonality, or certain type and level of noise. The data was transformed into a sample of 1000 instances for each of the original \glspl{Time-series} while it was unchanged for the number of data-points smaller than 1000. 

\cite{Soares2009Sep} generated 160 artificial datasets to obtain a wide range of cluster structures. There were two methods used to generate datasets: 1) a standard cluster model using Gaussian multivariate normal distributions, and 2) Ellipsoid cluster generator. There were three parameters selected for both techniques including i) the number of clusters which were the same for both cases (2, 4, 8, 16), ii) dimensions (2, 20 for Gaussian, and 50, 100 for Ellipsoid), and iii) the size of each cluster for both techniques were the same (uniformity in [10, 100] for 2 and 4 clusters case and [5, 50] for 8 and 16 clusters case). For each of the 8 combinations of cluster number and dimension, 10 different instances were generated, giving 80 datasets in each method.

\cite{Duch2011} used two artificially generated datasets out of a total of seven whereas the remaining five were the real-world problems. One artificially generated dataset had binary features, named as \emph{Parity}, whereas the other one with nominal features was known as \emph{Monks}. These support features are computed using \gls{QPC} projection.

\cite{ReifSep2012a} presented a novel data generator approach for numerical features and classification datasets that could be used as input datasets for \gls{Meta-learning} which represented an entirely different approach from \cite{Soares2009Apr}. The proposed system was able to generate datasets using genetic programming with customized parameters. In the proposed setting \gls{Meta-learning} could be supported in two different ways: 1) the \glspl{Meta-features} space could be filled in a more controlled way and the discovered "empty areas" could be populated rather than generating random datasets, and 2) thoroughly investigating \glspl{Meta-features} based on their descriptive power which could be useful for certain \gls{Meta-learning} problems and generating datasets with \glspl{Meta-features} allowed more controlled experiments that might lead to a significant utilization of particular \glspl{Meta-features}. Since the dataset was generating multiple \glspl{Meta-features} therefore this task was treated as a multi-objective optimization problem. The proposed system was able to incorporate a variable set of arbitrary \glspl{Meta-features}. The user was able to build a custom set of \glspl{Meta-features} simply by providing the functions that compute the \glspl{Meta-features}. 

\subsubsection{Datasets from Published Research} 
Another source of \gls{EoD} are the published \gls{Machine Learning} studies. As \gls{Machine Learning} has been one of the most active research areas in the last few decades where several experiments have been conducted, these experiments become a very useful way of gathering \gls{EoD} representing various domains. The additional benefit that usually comes with most of the datasets used in existing ML benchmarking and experimental studies is the relative ranking and predictive performance data for the evaluated ML algorithms. It is of particular interest as the ML algorithms performance data is used and needed as a target variable in the context of an \gls{Meta-learning} system. It is very time, memory and processor consuming task to compute performance measures for the massively large number of datasets and numerous configurations of learning algorithms. 

Usually, presumably due to space limitations on publications, researches publish only the final results with minimal details. However, in the context of \gls{Meta-learning}, relying on such minimal information leads to several problems, for example, in most of the instances researches only report the best algorithm, usually report limited number and detail of experimentations, mostly skip detailed configurations of the algorithms, etc. \cite{Vanschoren2014} introduced a novel platform for \gls{Machine Learning} research known as OpenML. \gls{Machine Learning} researchers can share datasets, algorithms, their configurations, and experiment setups on this platform which other researchers can use to compare results. OpenML framework is one of the possible solutions for most of the mentioned concerns which resolves two key challenges of \gls{Meta-learning} systems: i) gathering a large number of datasets from different domains, and ii) performances of the tested ML algorithms on these datasets. 

\subsubsection{Discussion and Summary}
An \gls{Machine Learning} system relies on a good training dataset to build a reliable and well-performing predictive model. Similarly, at the Meta-level, the \gls{Meta-knowledge} dataset is used as a training-set of \gls{Meta-learning}, and the quality of this \gls{Meta-knowledge} dataset is dependent on sufficient number and quality of \gls{EoD} from different domains. These \gls{EoD} are used to generate \glspl{Meta-features} which act as predictors and the estimated predictive performance evaluated ML algorithms for these \gls{EoD} are used as the target variable in the \gls{Meta-knowledge} dataset. However, gathering a sufficient number of real-world datasets is quite difficult. The real-world datasets which are used in various studies for experimentations are listed in Table~\ref{table:ExamplesofDatasets}. Most of the studies gathered datasets from the \gls{UCI} with different filtering options and the remaining few studies gathered datasets from different data-mining competitions. In most cases, the number of \gls{EoD} that are used to build \gls{Meta-knowledge} has been very low. However, as identified and shown in Table~\ref{table:SourcesofDatasets} there are various sources from which a relatively large (and quickly growing) number of real-world datasets representing different domains could be beneficially used in the future though they have not been used so far.

Some \gls{Meta-learning} researches resolved the problem of the number and quality of available datasets by building their \gls{Meta-knowledge} datasets using artificially generated \gls{EoD}. They have adopted two different approaches to generate these synthetic datasets: 1) by transforming real-world datasets; and 2) by utilizing statistical and genetic programming approaches. \cite{Bensusan2000b}, \cite{Pfahringer2000}, \cite{Soares2009Apr} and \cite{Wang2009} proposed different feature transformation approaches to generate different combinations of datasets from the limited number of real-world datasets. The statistical and genetic programming approaches were proposed by \cite{Soares2009Sep} and \cite{Duch2011} for \gls{Meta-learning} systems. In some of the approaches, statistical functions with a threshold (or cut-off) values are used to generate data while others used optimization techniques. \cite{ReifSep2012a} proposed an intelligent technique which does not generate random data, but fill the \glspl{Meta-features} in a more controlled way by discovering and populating the empty areas within the real-world datasets.

\begin{figure}[h!] 
    \centering
	\includegraphics[scale = 0.85, center]{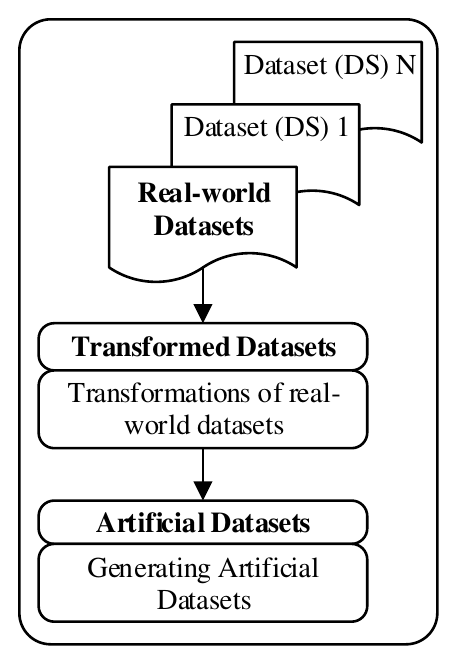} 
	\caption{Phase-wise collection of Examples of Datasets}
	\label{fig:FigRepositoryOfDatasets}
\end{figure}

Combining all the proposed approaches iteratively could offer a potential solution to the dataset scarcity; i.e., initially gathering the existing available real-world problems, then transforming those datasets by generating several others and finally applying various other techniques to generate artificial datasets independently (see Figure~\ref{fig:FigRepositoryOfDatasets}). Although this solution could be useful if the purpose would be only gathering a large number of \gls{EoD}, in the context of the MLL research the predictive performance data for numerous learning algorithms and their configurations are needed and not normally readily available. Considering all three necessary components of an \gls{Meta-learning} system, gathering datasets from published experimental evaluations and benchmarking of ML algorithms would seem to be more attractive, however, there are a lot of challenges with such data related to reporting only the best learning algorithms, publishing limited information of experimentations, availability of datasets used in the research, lack of detailed configurations of evaluated learning algorithms, etc. OpenML platform has attempted to address most of these issues focusing on the consistency and completeness of the gathered information but as it is in a preliminary stage of development it currently lacks a sufficiently large number of problems from different domains and sufficiently robust and comprehensive number of \gls{Machine Learning} algorithms tested for each of the datasets to be very useful in its current form.

\subsection{Meta-features Generation and Selection}
One of the primary applications of \gls{Meta-learning} is to recommend the best learning algorithm or to rank various ML algorithms for a new problem without the need for executing and evaluating these learning algorithms on the problem at hand. The role of such systems is to identify previously solved similar problems, and with the assumption that the previously found best algorithms will also work best on the new problem, make appropriate recommendations. As directly comparing large and complex datasets is normally infeasible, the similarity between different problems/datasets is carried out using a number of so called \glspl{Meta-features} offering a simplified representation of the problems/datasets. There are three most commonly used \gls{Meta-features} generation approaches which allow to induce a mapping between the characteristics of a problem and the best performing learning algorithms for the problem. These approaches are discussed in the following sections. 

\subsubsection{Descriptive, Statistical and Information-Theoretic Approach}
The \gls{DSIT} approach is the simplest and the most commonly used \gls{Meta-features} generation approach that extracts a number of \gls{DSIT} based \gls{Meta-features} values directly from a dataset representing an ML problem. The \gls{DSIT} based \glspl{Meta-features} and the related MLL approaches primarily based on such \glspl{Meta-features} are reviewed below.

\cite{Rendell1987} proposed \gls{VBMS} that was one of the earliest efforts towards data characterization. Only two descriptive \glspl{Meta-features}, namely: the number of training instances and the number of features, were used to select the best among three symbolic learning algorithms. Later \cite{Rendell1990} enhanced the existing system by adding useful \glspl{Meta-features} of complexity based on shape, size and structure. \gls{StatLog} project by \cite{King1995} further extended \gls{VBMS} \glspl{Meta-features} by considering a larger number of dataset characteristics. A problem was described in the context of its descriptive and statistical properties. Several characteristics of a problem spanning from simple (descriptive) to more complex (statistical) ones were generated and later used by various studies. These characteristics were used to investigate why certain algorithms perform better on a particular problem as well as to produce a thorough empirical analysis of the learning algorithms. 

\cite{Sohn1999} initially used most of the datasets and \glspl{Meta-features} that were used in \gls{StatLog} project which were later on enhanced with information-theoretic \glspl{Meta-features}. Furthermore, three new descriptive features were added by transforming the existing \glspl{Meta-features}, for example in the form of ratios. These \glspl{Meta-features} were used to rank several classification algorithms with considerably better performance as compared to the previous studies. It was also claimed that the classification error and execution-time are important response variables to choose a suitable classification algorithm for a problem. 

In the same year \cite{Lindner1999} proposed an extensive list of \gls{DSIT} based \glspl{Meta-features} of a problem under the name of \gls{DCT}. The authors distinguished three categories of dataset characteristics, namely simple, statistical and information-theory based measures. The descriptive \glspl{Meta-features} have been used to extract general characteristics of the dataset, whereas statistical characteristics were mainly extracted from numeric attributes, while information-theoretic based measures from nominal attributes. A \gls{CBR} approach to select the most suitable algorithm for a given problem was also proposed. 

\cite{ReifSep2012b} presented a novel approach of generating informative \glspl{Meta-features} by simply averaging overall attributes of the source datasets. They proposed a two-fold approach. In the first fold \gls{DSIT} based \glspl{Meta-features} are generated using the previously introduced traditional approach. The second fold is used to describe the differences over datasets that are not accessible using the typically used mean of \glspl{Meta-features} that have been computed in the first fold. This approach preserves more information on such \glspl{Meta-features} while producing a feature vector with a fixed size. An additional level of \glspl{Meta-features} selection is proposed to automatically select the most useful \glspl{Meta-features} out of the initially generated ones. All \glspl{Meta-features} that are used in the above studies are shown in Figure~\ref{fig:FigMetaFeatures}. 

\subsubsection{Landmarking Approach}
Another technique of \gls{Meta-features} generation is Landmarking which characterizes a dataset using the performance of a set of simple learners. Its main goal is to identify areas in the input space where each of the simple learners can be regarded as an expert \cite{Vilalta2002}. 

The basic idea behind Landmarking is to use the estimated performance of a learning algorithm on a given task for discovering additional information about its nature. A landmark learner or landmarker is defined as the learning mechanism whose performance is used to describe a problem \cite{Bensusan2000b}. Landmarkers posses a key property that their execution time is always shorter than the Base-learner's time, otherwise this approach would bring no benefit. In the remaining parts of this section, various studies dealing with Landmarking approaches are discussed in detail. 

One of the earliest studies on Landmarking was conducted by \cite{Bensusan2000b}. This approach is claimed to be simpler, more intuitive and effective than the \gls{DSIT} measures. A set of 7 landmarkers were trained on 10 different sets of equal size. Each dataset was then described by a vector of \glspl{Meta-features} (see Landmarkers branch of Figure~\ref{fig:FigMetaFeatures}), which are the error rates of the 7 landmarkers, and labelled by the target learners (see Table~\ref{table:TargetBaseLearners}) which produce the highest accuracy. Several experimentations have been performed to compare the landmarking approach with \gls{DSIT}. In the first experiment Landmarking was compared with 6 information-theoretic \gls{DCT} features of \cite{Lindner1999} (see information-theoretic \glspl{Meta-features} section of Figure~\ref{fig:FigMetaFeatures}). In most of the cases of this experiment landmarking outperformed the \gls{DSIT} based approach. In another experiment, the ability of landmarking to describe a problem and discriminate between two areas of expertise are highlighted. In most of the cases \gls{C5.0boost} \cite{Quinlan1998} landmarker performed best. The last experiment benchmarked 16 real-world datasets from the \gls{UCI} \cite{Bache2013} and the DaimlerChrysler where again the landmarking approach resulted in the best overall performance.

\cite{Pfahringer2000} also evaluated a landmarking approach while comparing it with the \gls{DSIT} \gls{Meta-features} generation approach - \gls{DCT}. They performed three types of experiments, namely: 1) Artificial rule list and sets generation, 2) Selecting learning models, and 3) Comparing the landmarking with the information-theoretic approach. These experiments were almost the same as performed by \cite{Bensusan2000b}, and the target learners (see Table~\ref{table:TargetBaseLearners}) were the same as used in \gls{METAL} project. In the first experiment the set of landmarkers consisted of a \gls{LDA}, Naive Bayes and \gls{C5.0trees} learners, while the base-learners performance relative to each other was predicted using \gls{C5.0boost}, \gls{LDA}, and \gls{Ripper}. In addition to 3 landmarkers, 5 descriptive \glspl{Meta-features} (shown in the descriptive approach in Figure~\ref{fig:FigMetaFeatures}) have also been extracted from 216 datasets. The Ripper was found to be the top performer in this experimentation. For selecting the best learning model experiment, the authors tried to investigate the capability of landmarking in deciding whether a learner involving multiple learning algorithms performs better than the other candidate algorithms. Here only \gls{C4.5} was used as a Meta-learner trained with 222 artificial boolean datasets and tested with 18 \gls{UCI} problems \cite{Bache2013}. Even though the landmarking accuracy was higher it did not have a significant effect on the overall performance of a system whose ultimate goal is to accurately select the best learning model. In the last experiment, the landmarking approach was compared with the \gls{DSIT} and also the combination of both approaches. 320 artificially generated binary datasets were produced where the combined approach performed best for all 10 Meta-learners followed by the landmarking with a significant difference as compared to \gls{DCT} approach. 

\cite{Soares2001} sample-based landmarkers used estimates of the performance of algorithms on a small sample of the data and then had been used as the predictors of the performance of those algorithms on the entire dataset. Additionally, a relative landmarker addressed the inability of the earlier landmarker to assess the relative performance of algorithms. This sampling-based relative landmarking approach was later compared with the \gls{DSIT} \gls{DCT} \glspl{Meta-features} \cite{Lindner1999} as done by most of the landmarking studies. The ten algorithms, listed in Table~\ref{table:TargetBaseLearners}, were used on 45 datasets, with more than 1000 instances, mostly gathered from the \gls{UCI} \cite{Bache2013} and the DaimlerChrysler repositories. These datasets have been ranked by the \emph{Nearest-Neighbour} using \gls{ARR} measure. To observe the performance of the ranking method, the authors varied the value of \emph{k} from 1 to 25. In comparison with other studies reported in the literature, the sample-based relative landmarking approach showed improvements in the algorithm ranking task as compared with the traditional \gls{DCT} measures.

\cite{Kopf2002} proposed a new approach for assessing the quality of case bases constructed using landmarking and DCT based \glspl{Meta-features}. The meta-learner was based on a case-base reasoning approach using the quality assessed cases. Tasks were described by their similarity, consistency, incoherency, uniqueness and minimality. A brief overview of the necessary requirements for the implementation of the case-based properties has also been provided in their study. A comprehensive experimentation was performed to compare variants of \gls{DCT} \gls{DSIT} approach, landmarking and their combinations. \glspl{Meta-features} were constructed for the experiments from the \gls{UCI} datasets (see Table~\ref{table:ExamplesofDatasets}) which contained up to 25\% missing values. Error rates for ten different classification algorithms from the METAL project were determined for different subsets of data characteristics mentioned in Table~\ref{table:TargetBaseLearners} and restricted to three Base-learners that are shown in Figure~\ref{fig:FigMetaFeatures}. The empirical results show the proposed approach in combination with \gls{DSIT}, and landmarking approaches as a promising one though not significantly different from previous meta-learning studies.

\cite{Abdelmessih2010} presented an overview of a landmarking operator and its evaluation. This landmarking operator was developed as part of an open-source RapidMiner data-mining tool. As mentioned repeatedly in the above studies, landmarkers selection is a critical process and the two basic criteria to select a landmarker were suggested in this study to be: 1) a landmarker has to be simple and require minimum execution (processing) time; and 2) it has to be simpler than the target learner(s). Following these conditions, RapidMiner provided the landmarkers shown in Figure~\ref{fig:FigMetaFeatures} and the target algorithms, for which the accuracy was predicted (see Table~\ref{table:TargetBaseLearners}). For the evaluation of these landmarkers, 90 datasets from the \gls{UCI} \cite{Bache2013} and other sources were collected with at least 100 samples in each. By following the existing studies, the landmarking operator has been compared with the \gls{DSIT} \glspl{Meta-features} of \gls{StatLog} \cite{King1995} and \gls{DCT} \cite{Lindner1999}, where landmarking resulted in \emph{5.1-8.3\%} overall performance improvement in all cases.

\begin{center}
{\small
\begin{longtable}{l|c|c}
\caption{Target Learners used in various studies}
\label{table:TargetBaseLearners} \\
\toprule
Target Learners & \cite{Bensusan2000b}, \cite{Pfahringer2000}, \cite{Soares2001}, \cite{Kopf2002}, \cite{Giraud-Carrier2005} &  \cite{Abdelmessih2010} \\
\midrule
\gls{C5.0trees} & \tick & \tick \\
\gls{C5.0rules} & \tick & - \\
\gls{C5.0boost} & \tick & - \\
\gls{NB} & \tick & \tick \\
\gls{IBL} & \tick & - \\
\gls{MLP} & \tick & \tick \\
\gls{RBF} & \tick & - \\
\gls{LDA} & \tick & - \\
\gls{Ripper} & \tick & - \\
\gls{Ltree} & \tick & - \\
\gls{k-NN} & - & \tick \\
\gls{RF} & - & \tick \\
\gls{OneR} & - & \tick \\ 
\gls{SVM} & - & \tick \\ \hline

Total Target Learners & 10 & 7 \\
\bottomrule
\end{longtable}}
\end{center}

\subsubsection{Model-based Approach}
Model-based \gls{Meta-features} generation is another effort towards task characterization in \gls{Meta-learning} domain. In this approach, the dataset is represented in a data structure that can incorporate the complexity and performance of the induced hypothesis. Later the representation can serve as a basis to explain the reasons behind the performance of the learning algorithm \cite{Giraud-Carrier2008}. Several research works utilizing the Model-based approach are discussed below.

\begin{figure}[ht!] 
    \centering
	\includegraphics[viewport=22 115 600 710, clip, scale = 0.88]{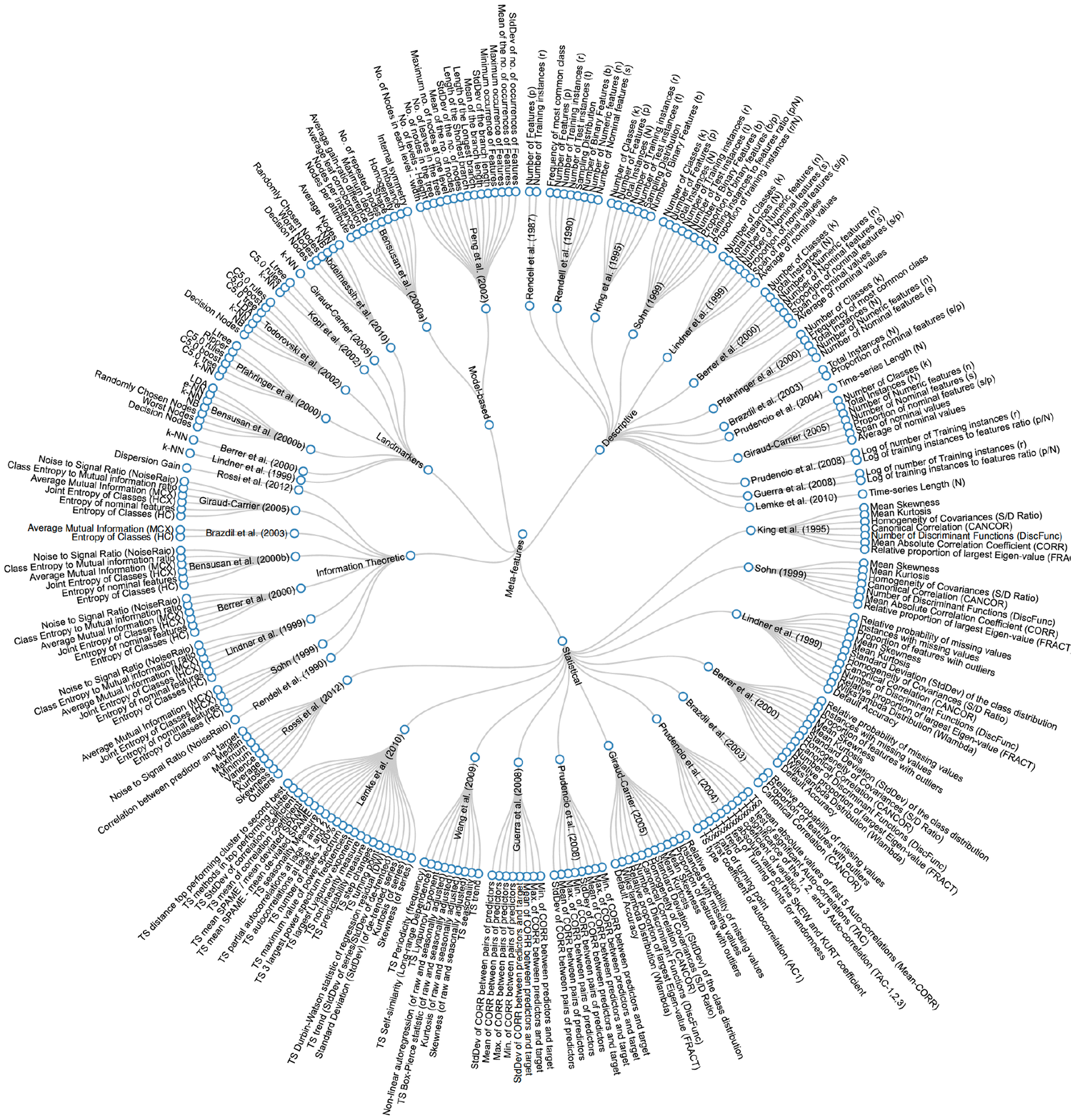} 
	\caption[The LOF caption]{Meta-features used in various studies\footnotemark}
	\label{fig:FigMetaFeatures}
\end{figure}

\cite{Bensusan2000a} study was an initial effort towards a model-based approach. The authors proposed to capture the information directly from the induced decision trees for characterizing the learning complexity. Figure~\ref{fig:FigMetaFeatures} lists the 10 descriptors computed from induced decision trees. Using these \glspl{Meta-features}, a task representation and algorithm to store and compare two different tree structures has been explained in detail with examples. The authors also elaborated on the motivation of using the induced decision trees directly rather than the predefined properties used in decision tree-based \glspl{Meta-features} that made explicit properties implicit in the tree structure. Finally, higher-order \gls{Meta-learning} approach was generalized by proposing data structures to characterize other algorithms. A tree-like structure was used for \gls{DT} in this work, sets were proposed for \emph{rule sets} and graphs for \glspl{NN}.

\cite{Peng2002} effort was towards improving the dataset characterization by capturing the structural shape and size of the decision tree induced from the dataset. For that purpose 15 features were proposed, known as \emph{DecT} and shown in Figure~\ref{fig:FigMetaFeatures}, which do not overlap with \cite{Bensusan2000a}. These measures were used to rank 10 learning algorithms in various experiments. In the first experiment \gls{DCT} \cite{Lindner1999} \gls{DSIT} \glspl{Meta-features} and 5 landmarkers (Worst Nodes Learner, Average Nodes Learner, \gls{NB}, and \gls{LDA}) were compared with DecT. The results proved the performance enhancement of the proposed approach. In another experiment, DecT measures were compared with the same \gls{DCT} measures and landmarkers to rank the learning algorithms based on the accuracy and time where again DecT performed better. The last experiment was performed to select \glspl{Meta-features} by reducing the number of features to 25, 15 and 8 respectively. The k-Nearest Neighbour algorithm, with various values of \emph{k} between 1 to 40, was used to select k datasets for ranking the performance of learning algorithms. The results suggested that the proposed feature selection did not significantly influence the performance of either DecT or even \gls{DCT}. Overall, DecT outperformed the other approaches. 

Neuro-cognitive inspired mechanism was proposed by \cite{Duch2011} to analyse learning-based transformations that generate useful hidden features for \gls{Meta-learning}. The types of transformations include restricted random projections, optimization using projection pursuit methods, similarity and general kernel-based features, conditionally defined features, and features derived from partial successes of various learning algorithms. The binary features were extracted from \gls{DT} and rule-based algorithms, continuous features were discovered by projection pursuit, linear \gls{SVM} and simple projections. \gls{NB} was used to calculate posterior probabilities along these lines while \gls{k-NN} and kernel methods were used to find similarity-based features. The proposed approach also evaluated and illustrated \gls{MDS} mappings and \gls{PCA}, \gls{ICA}, \gls{QPC}, \gls{SVM} projections in the original, one-, and two-dimensional space. Various real-world and synthetic datasets (details can be found in Table~\ref{table:ExamplesofDatasets}) were used for visualization and to analyse the kind of structures they create. The classification accuracies for each dataset were predicted using five classifiers including \gls{NB}, \gls{k-NN}, Separability Split Value Tree (SSV), Linear and Gaussian kernel \gls{SVM} in the original, one- and two-dimensional spaces. The results showed an overall significant improvement almost in all five algorithms as compared to the existing approach also proposed by the authors.

\footnotetext{Tabular representation of the visualization can be seen in Appendix~\ref{sec:appendix1})} 

\subsubsection{Discussion and Summary}
There are three common \gls{Meta-features} generation approaches proposed in the reviewed publications for \gls{Meta-learning}: 1) \gls{DSIT}, 2) Landmarking, and 3) Model-based. The \gls{DSIT} \glspl{Meta-features} approach was introduced at the early stage of \gls{Meta-learning} development where \cite{Rendell1987} proposed two descriptive features for \gls{VBMS}. Later on \cite{Rendell1990} added more descriptive features to the original list. The statistical \glspl{Meta-features} were introduced by \cite{King1995}, and \cite{Sohn1999} proposed information-theoretic features combined with some existing descriptives to represent a problem at a Meta-level. Finally, \cite{Lindner1999} proposed an extensive list of \gls{DSIT} \glspl{Meta-features}, known as \gls{DCT}. The \gls{DCT} measures became a benchmarked approach to represent a problem using the \gls{DSIT} approach. These measures were later used in several studies for experimentation, e.g. \cite{Berrer2000}, \cite{Giraud-Carrier2005}, etc., and compared with other \gls{Meta-features} approaches.

Landmarking and Model-based approaches are more recent ones and have been outperforming the \gls{DSIT} in almost all the comparative studies. The earliest study on landmarking was conducted by \cite{Bensusan2000b} where the approach was claimed to be simpler, more intuitive and efficient than \gls{DSIT}. The proposed approach was compared with and outperformed information-theoretic measures of \gls{DCT} with a significant difference. Though one common deficiency that is observed in several \gls{Meta-learning} studies is the use of a smaller number of \gls{EoD} for experimentations which raised a question on the significance of the reported results. \cite{Pfahringer2000} used a different set of landmarkers but the same target learners as \cite{Bensusan2000b}. This work can be considered to offer improvements to the previous one in two aspects: 1) a huge number of synthetic datasets were used, and 2) some descriptive \glspl{Meta-features} were combined with the landmarkers. This approach was also compared with \gls{DCT} features where landmarking showed significant improvement in the results. Similarly \cite{Soares2001}, \cite{Kopf2002} and \cite{Abdelmessih2010} used different sets of target learners, landmarkers, number of dataset examples, and compared their approaches with a different set of \gls{DSIT} measures. All of them reported improved results of the landmarking approach over the \gls{DSIT}.

\cite{Bensusan2000a} approach to characterizing the learning complexity by directly inducing MFs from the model is the earliest work towards model-based approach. In this work, 10 descriptors (MFs) were computed from the induced decision trees which can be seen in Figure~\ref{fig:FigMetaFeatures}. \cite{Peng2002} effort was towards improving this characterization by focusing on the structural shape and size of the decision tree induced from the datasets. The other dimension of this work was to compare the proposed model-based approach with \gls{DCT}, \gls{DSIT} and landmarking measures. Various experimentations were performed with variations of \glspl{Meta-features} and landmarkers where the model-based approach consistently performed better. A problem with these Meta-level problem representations is that they can not easily accommodate non-stationary environments. Most of the effort has been dedicated to the stationary environment, even though some recent studies are addressing \glspl{Meta-features} for a dynamically changing environment, i.e. \cite{Rossi2014}, but these are not mature enough to represent the entire domain. Although \cite{Rossi2014} used traditional \gls{Meta-features} that are used to characterize stationary data, only those \glspl{Meta-features} were computed that characterize individual variables. Moreover, there are separate features computed for training and selection windows. Their reliability is highly dependant on the number and quality of examples, thus the larger the number of examples in a window, the potentially higher the reliability of the problem representation at the Meta-level. However, in a rapidly changing environment, there is often a very limited number of examples between consecutive concept changes. Hence there is an unaddressed need for novel \glspl{Meta-features} and approaches that can cope with small data samples. 

\begin{figure}[h!] 
    \centering
	\includegraphics[scale = 0.81, center]{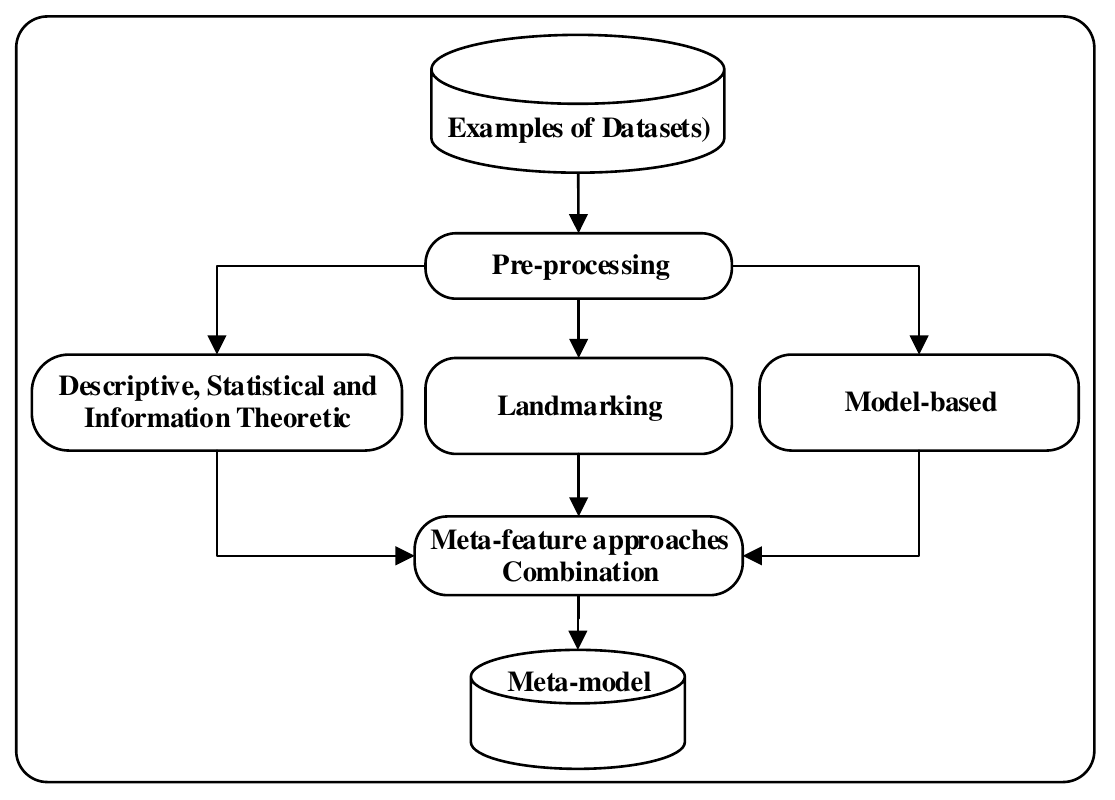} 
	\caption{Combining Significant Meta-features from various approaches}
	\label{fig:FigMetaFeaturesCombination}
\end{figure}

From the above studies, it can be observed that combining significant \glspl{Meta-features} from different feature generation approaches might be useful as shown in Figure~\ref{fig:FigMetaFeaturesCombination}. 

\subsection{Base-level Learning}
In the context of \gls{Meta-learning}, \gls{Base-learning} algorithms are used to build predictive models on input datasets and for \gls{Meta-learning} purposes are used to compute a set of performance measures, i.e, accuracy, execution-time, etc. These performance measures are combined with their respective \glspl{Meta-features} in the \gls{Meta-knowledge} database. A Meta-learner uses these performances as a target variable. The remaining sections discuss several studies concerned with the roles and characteristics of individual and combined \gls{Base-learning} algorithms utilised within the \gls{Meta-learning} context.

\cite{Brazdil2003} proposed an \gls{Meta-learning} based approach to rank candidate algorithms where \gls{k-NN} was used to identify the datasets that were most similar to the query dataset. The pool of candidate algorithms contained an ensemble method, namely \gls{C5.0boost}, which performed well for 19 out of 53 datasets in the presence of 9 other algorithms. The performance of ensemble methods was ranked with individual learning algorithms. In general, several kinds of research used \gls{C5.0boost} ensemble method with other individual algorithms and found it to be the top-performing method. 

The applicability of \gls{Meta-learning} on a \gls{Time-series} task was demonstrated by \cite{LemkeJun2010}. Several individuals and a combination of forecasting algorithms were used to investigate which model works best in which situation. In the experiments 5 forecasting combination methods were used including 1) \emph{simple average} where all available forecasts are averaged, 2) \emph{simple average with trimming} which do not take the worst-performing 20\% models into account, 3) \emph{variance-based method} where weights for a linear combination of forecasts are determined using past forecasting performance, 4) \emph{out-performance method} which determines weights based on the number of times a method performed best in the past, and 5) \emph{variance-based pooling} which first groups past forecast performance into 2-3 clusters and then takes their average to obtain the final forecast. The results of these experiments showed that the forecast combination methods perform better than individual models which are listed in Table~\ref{table:BaseLearningStrategy}. Further discussion of this work can be found in Chapter~\ref{sec:metaLearning}.

\cite{Menahem2011} proposed a new \gls{Meta-learning} based ensemble scheme for one-class problems know as TUPSO. The TUPSO combined one-class Base-classifiers via a Meta-classifier to produce a single prediction. The \gls{Base-learning} component generates predictions of classifiers that are used to extract aggregated \glspl{Meta-features} as well as one-class accuracy and f-score estimates. The one-class performance evaluator computed each Base-classifier on only positively labelled instances using 4 algorithms including 1) global density estimation, 2) peer group analysis, 3) \gls{SVM}, and 4) attribute distribution function approximation (ADIFA) on 53 distinct datasets (details can be seen in Table~\ref{table:ExamplesofDatasets}). There are 15 aggregated \glspl{Meta-features} computed from the predictions of Base-classifiers that are clustered into four groups: 1) summation-based (votes, predictions, weighted predictions, power and log of weighted predictions), 2) variance-based (votes, predictions, and weighted), 3) histogram-based, and 4) representation-length. In an empirical evaluation an ensemble method, Fixed-rule, produced worse classification accuracy when compared to \gls{Meta-learning} based ensembles - TUPSO.

\begin{center}
{\small
\begin{longtable}{c|p{2.7cm}|p{7.2cm}|p{2.8cm}}
\caption{Base-level learning strategies used in different studies}
\label{table:BaseLearningStrategy} \\
\toprule
Research Work & Sampling Strategy & Base-learners & Performance Measure \\
\midrule
\cite{King1995} & 9-fold \gls{CV} for datasets with less than 2500 instances & \gls{k-NN}, \gls{RBF}, Density Estimation, \gls{CART}, \gls{INDCART}, Back-propagation, NewID, \gls{C4.5}, \gls{CN2}, \gls{Quadra}, Cal5, AC\textsuperscript{2}, \gls{SMART}, Logistic Regression, \gls{FLD}, ITrule, \gls{CASTLE}, \gls{NB} & Misclassification error, Run-time speed \\

\cite{Bensusan2000b} & stratified 10-fold \gls{CV} & \gls{NB}, \gls{MLP}, \gls{RBF}, \gls{C5.0trees}, \gls{C5.0rules}, \gls{C5.0boost}, \gls{IBL}, \gls{LDA}, \gls{Ripper}, \gls{Ltree} & - \\

\cite{Pfahringer2000} & 10-fold \gls{CV} & \gls{NB}, \gls{MLP}, \gls{RBF}, \gls{C5.0trees}, \gls{C5.0rules}, \gls{C5.0boost}, \gls{IBL}, \gls{LDA}, \gls{Ripper}, \gls{Ltree} & \gls{MAE}  \\

\cite{Soares2001} & - & \gls{NB}, \gls{MLP}, \gls{RBF}, \gls{C5.0trees}, \gls{C5.0rules}, \gls{C5.0boost}, \gls{IBL}, \gls{LDA}, \gls{Ripper}, \gls{Ltree} & - \\

\cite{Peng2002} & 10-fold \gls{CV} & \gls{C5.0trees}, \gls{C5.0rules}, \gls{C5.0boost}, \gls{Ltree}, \gls{LDA}, \gls{NB}, \gls{IBL}, \gls{MLP}, \gls{RBF}, \gls{Ripper} & \gls{MSE}, Run-time speed \\

\cite{Todorovski2002} & 10-fold \gls{CV} & \gls{C5.0trees}, \gls{C5.0rules}, \gls{C5.0boost}, \gls{Ltree}, \gls{Ripper}, \gls{NB}, \gls{k-NN} \footnote{k=1}\addtocounter{footnote}{-1}\addtocounter{Hfootnote}{-1}, \gls{LDA} & \gls{MSE} and \gls{Spearman} \\

\cite{Kopf2002} & 10-fold \gls{CV} & \gls{NB}, \gls{MLP}, \gls{RBF}, \gls{C5.0trees}, \gls{C5.0rules}, \gls{C5.0boost}, \gls{IBL}, \gls{LDA}, \gls{Ripper}, \gls{Ltree} & - \\ 

\cite{Brazdil2003} & 10-fold \gls{CV} & \gls{C5.0trees}, \gls{C5.0rules}, \gls{C5.0boost}, \gls{Ltree}, \gls{IBL}, \gls{Ripper}, \gls{LDA}, \gls{NB}, \gls{MLP}, \gls{RBF} & \gls{ARR} \\ 

\cite{Prudencio2004} & I: Train and test and II: train, test and validate & I: J.48 and II: \gls{MLP} & \gls{MAE} \\

\cite{Giraud-Carrier2005} & 10-fold \gls{CV} & \gls{NB}, \gls{MLP}, \gls{RBF}, \gls{C5.0trees}, \gls{C5.0rules}, \gls{C5.0boost}, \gls{IBL}, \gls{LDA}, \gls{Ripper}, \gls{Ltree} & - \\ 

\cite{Guerra2008} & 10-fold \gls{CV} & \gls{MLP}\footnote{hidden nodes = 1, 2, 3, 8, 16, 32} & Normalized \gls{MSE} \\

\cite{Wang2009} & 80\% Training and 20\% testing partition & \gls{ES}, \gls{ARIMA}, \gls{RW}, \gls{NN} & - \\ 

\cite{kadlec2009architecture} & Leave-one-out \gls{CV} & \gls{MLR}, \gls{MLP}, \gls{RBF}, Lazy-learning & \gls{MSE} and \gls{Spearman} \\

\cite{LemkeJun2010} & 10-fold \gls{CV} & \gls{ARIMA}, Structural model, Iterated (single exponential smoothing, Taylor smoothing, theta, \gls{NN}, elman \gls{NN}), Direct (regression, theta \gls{MA}, single exponential smoothing, Taylor smoothing, \gls{NN}) & \gls{SMAPE} \\

\cite{Abdelmessih2010} & 10-fold \gls{CV} & \gls{NB}, \gls{k-NN}, \gls{MLP}, \gls{C5.0trees}, \gls{RF}, \gls{OneR}, \gls{SVM} & \gls{RMSE} \\

\cite{Rossi2012} & Training and testing & \gls{RF}, \gls{SVM}, \gls{CART}, \gls{PPR} & Normalized \gls{MSE} \\

\cite{Rossi2014} & Training and testing & \gls{RF}, \gls{SVM}, \gls{CART}, \gls{PPR}, \gls{MARS} & Normalized \gls{MSE} \\
\bottomrule
\end{longtable}}
\end{center}

\subsection{Discussion and Summary}
The \gls{Meta-knowledge} database usually consists of \glspl{Meta-features} and performance measures (target) of different learning algorithms which are predicted accuracies for \gls{EoD}. These predictive values are computed, in the context of \gls{Meta-learning}, through \gls{Base-learning}. The predictive accuracies of learning algorithms are used as a basis for identifying the best algorithm from the pool of methods, their ranking, and/or a combination. Another level of complexity is introduced by the different parametrizations of the algorithms which were overlooked by several studies where only default configurations were considered. Furthermore, most of them selected only the best algorithm from the pool to minimize the representation complexity of \gls{Meta-knowledge} dataset, therefore very few of them stored information about the ranking and relative performance of evaluated BLLs. Table~\ref{table:BaseLearningStrategy} shows different learning strategies, Base-learners and performance measures that various \gls{Meta-learning} studies used at the Base-level. It can be observed that the 10-fold cross-validation strategy, \gls{MAE} accuracy measure and few learning algorithms have become a norm to use at the Base-level. The same Base-level learning strategies are used in some \gls{Meta-learning} studies for \gls{Time-series} with different ARIMA and exponential smoothing algorithms. Another common deficiency that can be observed from various studies is related to the granularity of information that is being stored in \gls{Meta-knowledge} database.

Table~\ref{table:PerformanceMeasures} summarises and groups the reviewed studies according to the four dominant performance measures which were used as the target variable for an \gls{Meta-learning} system.

\begin{center}
{\small
\begin{longtable}{p{3.5cm}|p{6.4cm}|p{4.5cm}}
\caption{Different Performance Measures that are used in MLL studies}
\label{table:PerformanceMeasures} \\
\toprule
Performance Measure(s) & Description & Research Work \\
\midrule
Best learning algorithm	& The performance measure only contains of the classification accuracy of best learning algorithm for each single dataset & \cite{Utgoff1984}, \cite{Graner1994}, \cite{King1995}, \cite{Bensusan2000a} \\

Ranking of learning algorithms & To predict a ranked list of learning algorithms in a pool which are sorted based on a performance measure, e.g. classification accuracy, run-time, etc. & \cite{King1995}, \cite{Brazdil2003}, \cite{Vilalta2004} \\

Quantitative Prediction \cite{ReifFeb2012} & To directly predict the performance of the target learning algorithm in an appropriate unit, i.e., by training separate regression model for each target algorithm	& \cite{Gama1995}, \cite{Sohn1999}, \cite{Kopf2002}, \cite{Bensusan2001}, \cite{ReifFeb2012} \\

Predicting Parameters & The \gls{Meta-learning} target variable could be one parameter value or a set of values & \cite{Soares2004}, \cite{Soares2006}, \cite{kadlec2009architecture}, \cite{LemkeJun2010}  \\
\bottomrule
\end{longtable}}
\end{center}

\subsection{Meta-learning} \label{sec:metaLearning}
The \gls{Meta-knowledge} induced for the \gls{Meta-learning} purposes provides a means for making informed decisions in relation to which algorithms are likely to perform best/well for a given problem \cite{Giraud-Carrier2008}. This section presents the history of the most promising decision-support systems for algorithm selection, followed by a review of the applicability of \gls{Meta-learning} to the supervised and unsupervised learning algorithms.

\subsubsection{Existing Systems}
Based on the reviewed literature, \cite{Utgoff1984} can be considered as the earliest effort towards developing \gls{Meta-learning} systems where a system named \gls{STABB} was proposed. It was a demonstration that a learner's bias could be adjusted dynamically. Later this work became an initial point of reference and was enhanced in several studies. One of them was \glsreset{VBMS}\gls{VBMS} by \cite{Rendell1987}, where a relatively simple \gls{Meta-learning} system was proposed. \gls{VBMS} selected the best among three symbolic learning algorithms as a function of only two dataset characteristics, namely, the number of training instances and the number of features. As mentioned in one of the previous sections, this was then further improved in \cite{Rendell1990}.

\glsreset{MLT}\gls{MLT} project by \cite{Graner1994} was one of the initial attempts to address the applications of \gls{Meta-learning}. \gls{MLT} produced a toolbox consisting of 10 symbolic learning algorithms for classification. The part of \gls{MLT} project that assists with the algorithm selection is known as a Consultant. The Consultant was based on a stand-alone expert system that maintained a knowledge-base that considered the experiences acquired from the evaluation of learning algorithms. Considerable insight into many important \gls{Machine Learning} issues was gained which had been translated into rules that formed the basis of Consultant-2. Consultant-2 was also an expert system for algorithm selection that gathered user inputs through a set of questions about the data, the domain and user preferences. Based on the user response relevant rules led to either additional questions or, eventually, a classification algorithm recommendation. Although its knowledge base had been built through an expert-driven knowledge engineering rather than via \gls{Meta-learning} it still stands out as the first automatic tool that systematically related application domain and dataset characteristics to the most suitable classification algorithms. Additionally, Consultant-3 provided advice and help on the combination of learning algorithms. It is also able to perform self-experimentation to determine the effectiveness of an algorithm on a learning problem.

In \glsreset{StatLog}\gls{StatLog} project \cite{King1995} presented the results of comprehensive experiments on classification algorithms. The project was an extension of \gls{VBMS} by considering a larger number of \glspl{Meta-features}, together with a broad class of candidate models for algorithm selection. It aimed to compare several symbolic learning algorithms on twelve large real-world classification tasks. Some \gls{Meta-learning} algorithms were used for model selection tasks where statistical measures, e.g., skewness, kurtosis and covariance, that produced higher accuracy were reported. Additionally, a thorough empirical analysis of 16 classifiers on 12 large real-world datasets and learning models using accuracy and execution time measures of performance were produced. There is no single algorithm that performed best in the experimentation phase. Symbolic algorithms resulted in the best performance for datasets with extreme distributions, i.e., where distribution was far from normal (i.e., specifically with skew $>$ 1 and kurtosis $>$ 7), and the worst in the scenarios where the datasets were evenly distributed. In contrast, the Nearest Neighbour algorithm was found to be accurate for datasets containing evenly distributed in terms of scale and importance of the features.

The \glsreset{METAL}\gls{METAL} project was developed to facilitate the selection of the best-suited classification algorithm for a data-mining task \cite{Berrer2000}. It guides the user in two ways: 1) in discovering new and relevant \glspl{Meta-features}; and 2) in a selection or ranking of classifiers using an \gls{Meta-learning} process. The main deliverable of this project was the \glsreset{DMA}\gls{DMA}, a Web-based \gls{Meta-learning} system for the automatic selection of classification learning algorithms \cite{Giraud-Carrier2005}. The \gls{DMA} returned a list of ten algorithms that were ranked according to how well they met the stated goals in terms of accuracy and training time. It implemented ranking mechanisms by exploiting the ratio of accuracy and training time. The choice of an algorithm ranking, rather than selecting the best-in-class, was motivated by a desire to give as much information as possible and as a consequence, a number of algorithms could be subsequently executed on the dataset. 

The \glsreset{METALA}\gls{METALA}, developed by \cite{Botia2001}, was an agent-based architecture for distributed Data Mining, supported by \gls{Meta-learning}. The system supported an arbitrary number of algorithms and tasks, and automatically selected an algorithm that appeared best from the pool of available algorithms. Like in the case of \gls{DMA}, each task was characterized by \gls{DSIT} features relevant to its usage, including the type of input data it required, the type of model it induced, and how well it handled noise. It had been designed to automatically carry out experiments with each learner and task, and induce a Meta-model for an algorithm selection. As new tasks and learning algorithms were added to the system, corresponding experiments were performed and the Meta-model was updated.

The \glsreset{IDA}\gls{IDA} provided a \gls{KD} ontology that defined the existing techniques and their properties \cite{Bernstein2001}. It supported three algorithmic steps of the \gls{KD} process, including preprocessing, data modelling and post-processing. The approach used in this system was the systematic enumeration of valid data-mining processes so that potentially fruitful options were not overlooked, and effective ranking of these valid processes based on user-defined preferences e.g., prediction accuracy, execution speed, etc. \gls{IDA} systematically searched for an operation whose pre-conditions have been met and whose indicators were consistent with the user-defined preferences. Similarly, its post-conditions searched for an operation and the process terminated once the goal had been reached. Once all valid \gls{KD} processes had been generated, a heuristic ranker was applied to return user-specified goals. \cite{Bernstein2005} research had focused on extending the IDA approach by leveraging the interaction between ontologies to extract deep knowledge and case-based reasoning for \gls{Meta-learning}. The system also used procedural information in the form of rules fired by an expert system. The case-base was built around 53 features to describe cases and the ontology came from human experts. 

\cite{Mierswa2006} developed a landmarking operator in RapidMiner as part of \gls{PaREn} project, which was an open-source system for data mining. This operator extracted landmarking features from a given dataset by applying seven fast computable classifiers on it (shown in Figure~\ref{fig:FigMetaFeatures}). 

\begin{center}
{\small
\begin{longtable}{c|p{2.2cm}|p{2.0cm}|p{4.0cm}|p{4.0cm}}
\caption{Existing Meta-learning Systems}
\label{table:ExistingSystems} \\
\toprule
Research Work & Title & Approach & Contributions & Limitations \\
\midrule
\cite{Utgoff1984} & \gls{STABB} & Statistical & Initial effort towards \gls{Meta-learning} & Limited to altering only one kind of learner's bias with fixed order of choices \\
\cite{Rendell1987} & \gls{VBMS} & Descriptive & Biases are dynamically located and adjusted according to problem characteristics and prior experience & VBMS is a relatively simple \gls{Meta-learning} system that learns to select the best among three symbolic learning algorithms as a function of only two dataset characteristics \\
\cite{Rendell1990} & Empirical Learning as a Function of Concept Character & \gls{DSIT} & Complex \glspl{Meta-features} based on shape, size and concentration, and artificial data generation is used & These complex \glspl{Meta-features} are expensive to compute \\
\cite{Graner1994} & \gls{MLT} & Rule-based & An expert system for algorithm selection by gathering user input through questions and trigger relevant rules while the knowledge-base was built through expert-driven knowledge engineering & Its knowledge base was built through expert-driven knowledge engineering rather than \gls{Meta-learning} \\
\cite{King1995} & \gls{StatLog} & Statistical & A thorough empirical analysis of learning algorithms and models is produced by comparing several symbolic learning algorithms on twelve real-world classification tasks & For a given dataset, algorithms were characterized only as applicable or non-applicable, i.e., they do not provide a way to rank the algorithms; furthermore, that characterization was based on a simple comparison of accuracies regardless of any statistical significance test \\
\cite{Berrer2000, Giraud-Carrier2005} & \gls{METAL} - \gls{DMA} & \gls{DSIT} and Landmarking & Discovers new and relevant \glspl{Meta-features} and algorithm ranking in terms of accuracy and execution time & The outcome of the prediction model is only the best classifier for the new dataset. It does not support multi-operator workflows \\
\cite{Botia2001} & \gls{METALA} & Model-based & Agent-based architecture for distributed data-mining, automatically carry out experiments and induce a Meta-model for algorithm selection, it provides architectural mechanisms necessary to scale the \gls{DMA} & \gls{DMA}'s \glspl{Meta-features} are used to represent a problem, no contribution to introduce new features \\
\cite{Bernstein2001} & \gls{IDA} & Model-based & Its goal is to rank pre-processing, modelling and post-processing steps that are both valid and consistent with the user-defined preferences & The data should be already pre-processed considerably by the user for IDA to model it and evaluating the resulting models \\ 
\cite{Bernstein2005} & \gls{IDA} - An Ontology-based Approach & Model-based & Extending IDA approach by leveraging the interaction between ontology for deep knowledge and Case-Based Reasoning for \gls{Meta-learning} & The case-based is built on fixed 53 features and the system is still in the early stages of implementation \\ 
\cite{Mierswa2006} & \gls{PaREn} & Landmarking & A Landmarking operator for \gls{Meta-learning} developed in RapidMiner & Very limited \gls{EoD} (from \gls{UCI}) are used to build \gls{Meta-knowledge} \\
\cite{eLico2012} & \gls{e-LICO} & Model-based & An e-Laboratory for interdisciplinary collaborative research in data-mining and data-intensive science & Meta-learning component is using RapidMiner's landmarking system which is built on only 90 \gls{UCI} datasets \\
\bottomrule
\end{longtable}}
\end{center}

\begin{figure}[h!]
    \centering
	\includegraphics[scale = 0.85, center]{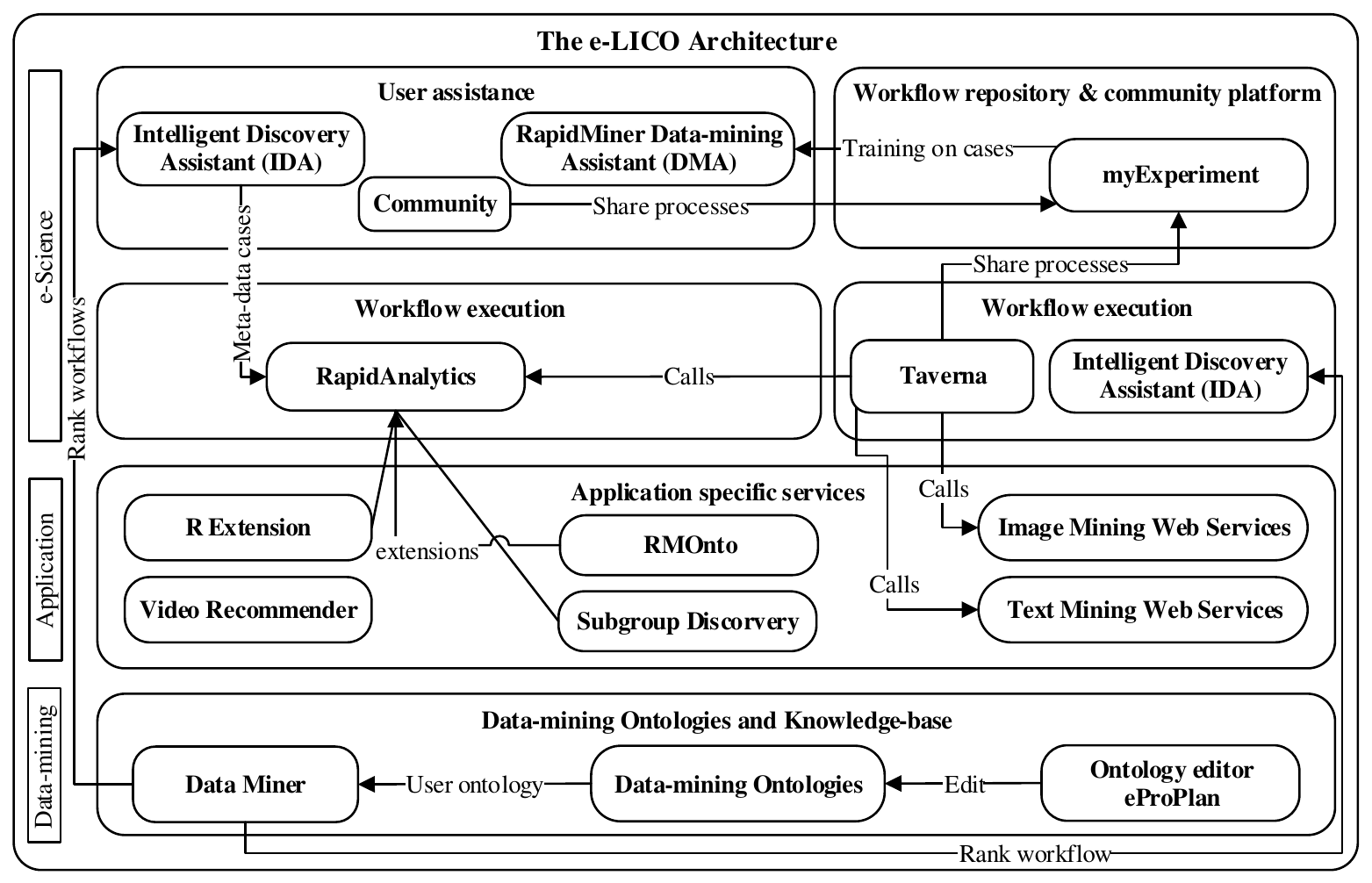} 
	\caption{e-LICO project architecture}
	\label{fig:ExistingSystemseLico}
\end{figure}

\gls{e-LICO} was a project for data-mining and data-intensive science \cite{eLico2012}. This project comprised of three layers: 1) e-Science, 2) Application, and 3) Data-mining. The e-Science and data-mining layers formed a generic environment that was adapted to different scientific domains by customizing the application layer. The architecture of \gls{e-LICO} project was shown in Figure~\ref{fig:ExistingSystemseLico}.

The e-Science layer was built on an open-source e-science infrastructure that supported content creation through collaboration at multiple scales in dynamic virtual communities. The Taverna\footnote{A suite of tools used to design and execute scientific workflows and experimentation. http://www.taverna.org.uk}, \gls{RapidAnalytics} and RapidMiner \cite{Mierswa2006} components had been used to design and enact data-analysis workflows. The system also provided a variety of general-purpose and application-specific services and a broad tool-kit in designing and sharing such workflows with data-miners all over the word using \emph{myExperiment} portal. The IDA \cite{Bernstein2001} exposed \gls{Meta-learning} capabilities by automatically creating processes tailored for the specification of input data and a modelling task. The RapidMiner's \gls{DMA} component helped to design processes by recommending operators that fitted well with the existing operators in a process. The data-mining layer provided comprehensive multimedia data-mining tools that were augmented with preprocessing and learning algorithms developed specifically to meet challenges of data-intensive, knowledge-rich sciences. The knowledge-driven data-mining assistant relied on a data-mining ontology and knowledge-base to propose ranked workflows for a given task. The application layer initially came as an empty shell that had to be built by the domain user from different components of the system. At the application layer, \gls{e-LICO} was showcased in two application domains: 1) a systems biology, and 2) a video recommendation task.

\subsubsection{Regression and Classification Problems}
This section covers various aspects of \gls{Meta-learning} that are used for regression and classification tasks in different systems.

\cite{Todorovski2002} addressed a novel approach of predictive clustering trees to rank classification algorithms using dataset properties. The approach was to illustrate \gls{Machine Learning} algorithms ranking where the relative performance of the algorithms had to be predicted from a given dataset's \glspl{Meta-features}. For that purpose the performance of eight Base-level algorithms, mentioned in Table~\ref{table:BaseLearningStrategy}, has been measured on 65 classification tasks gathered from the \gls{UCI} repository and the \gls{METAL} project. Furthermore, \gls{DSIT} dataset characteristics from \gls{StatLog} and \gls{DCT} were combined to create an \gls{Meta-knowledge} dataset consisting of 33 \glspl{Meta-features}. The properties of individual attributes were aggregated using average, minimum or maximum functions. The landmarking approach was used in this study with 7 simple and fast learners, shown in Figure~\ref{fig:FigMetaFeatures}, to investigate the ranking task performance. The proposed dataset characterization approach with clustering tree outperformed with a significant margin the \gls{DCT} and the histogram approach which used a grained aggregation of \gls{DCT} properties.

\cite{Vilalta2002} presented four approaches to \gls{Meta-learning} consisting of learning from Base-learners; 1) Stacked generalization, 2) Boosting, 3) Landmarking, and 4) Meta-decision trees. The information collected from the performance of \gls{Base-learning} algorithms were incorporated into the \gls{Meta-learning} process. Stacked generalization was considered a form of \gls{Meta-learning} where each set of Base-learners was trained on a dataset and the original feature representation was then extended with the predictions of the Base-learners. These predictions were received by successive layers as inputs and the output was passed on to the next layer. A single (Meta-)learner at the topmost layer computed the final prediction. Boosting was another approach that was considered as a form of \gls{Meta-learning}. It generated a set of Base-learners by generating variants of the training set using sampling with replacement technique under a weighted distribution. This distribution is modified for every new variant by assigning more weights to the incorrectly classified examples using the most recent hypothesis. Boosting took the predictions of each hypothesis over the original training set to progressively improve the classification of those examples for which the last hypothesis failed. 

In the last proposed approach, the Base-learners consisted of a combination of several inductive models induced from Meta-decision trees. A decision tree was built where each internal node represented a \gls{Meta-features} that predicted a class probability for a given example by a set of models whereas the leaf nodes corresponded to a predictive model. Given a new example, the Meta-decision tree selected the most suitable model to predict the target value. \cite{Todorovski2003} used the same approach for \gls{Meta-learning} discussed in this section. 

An instance-based learning algorithm, \gls{k-NN}, was used to identify the datasets that were most similar to the one at hand by \cite{Brazdil2003}. The candidate Base-learning algorithms were not ranked but selected based on a multi-criteria aggregated measure that took accuracy and time into account. The proposed methodology had been evaluated using various experiments and analysis at the Base- and Meta-level learning. The Meta-data used in this study was obtained from \gls{METAL} project which contained estimates of accuracy and time for 10 algorithms (listed in Table~\ref{table:BaseLearningStrategy}) on 53 datasets, using 10-fold \gls{CV}. The \gls{k-NN} algorithm was used at the Meta-level to select the best candidate algorithm for a new dataset. For two values of the number of neighbours, 1 and 5, the \gls{k-NN} showed a significant improvement in the results, particularly with k=1, as compared to the trial-and-error approach. 

Two \gls{Meta-learning} approaches were investigated to select models for \gls{Time-series} forecasting by \cite{Prudencio2004} in different case studies. In the first case study, a single \gls{Base-learning} algorithm was used to select models to forecast stationary \gls{Time-series}. The base-level and meta-level learning algorithms and configurations are given in Table~\ref{table:BaseLearningStrategy} and Table~\ref{table:MetaLearningStrategy} for both case studies while details of datasets and \glspl{Meta-features} are listed in Table~\ref{table:ExamplesofDatasets} and Figure~\ref{fig:FigMetaFeatures} respectively. In another case study, a more recent and sophisticated approach - NOEMON \cite{kadlec2009architecture} was used to rank three models of the M3-Competition. In both case studies, the experiments revealed significant results by taking into account the quality of algorithm selection and forecasting algorithm performance aspects of the selected models.

Active \gls{Meta-learning} method, in combination with \emph{Uncertainty Sampling} and outlier detection, had been proposed by \cite{Prudencio2008} to support the selection of informative and anomaly-free Meta-examples for \gls{Meta-learning}. Some experiments were performed in a case study where \gls{MLP} was used to predict the accuracies of 50 regression problems at the Base-level learning (the details can be seen in Table~\ref{table:ExamplesofDatasets}) and \gls{k-NN}\footnote{k = 1, 3, 5, 7, 9 and 11 nearest neighbours} at the Meta-level. The \glspl{Meta-features} used in the case study consisted of 10 simple and statistical measures which can be seen in Figure~\ref{fig:FigMetaFeatures}. The results of the experiments revealed that the proposed approach was significantly better than the previous work on Active \gls{Meta-learning}. Also, the \emph{Uncertainty Sampling} method increased the performance when the outliers were eliminated from the \gls{Meta-knowledge} which affected 5\% of the data.

\cite{Guerra2008} used \gls{SVM}, with different kernel functions, as a Meta-regressor to predict the performance of a candidate algorithm, \gls{MLP}, based on descriptive and statistical features of the learning tasks. For experimentation purposes, the input datasets and \glspl{Meta-features} used in this study were the same as those in the \cite{Prudencio2008} work. The \gls{MLP} was used as a base-learner to compute the normalized \gls{MSE} which was averaged over 10 training runs. Table~\ref{table:BaseLearningStrategy} contains details of the learning strategy which were used at the base-level. At the meta-level, \gls{SVM} with different kernel functions (listed in Table~\ref{table:MetaLearningStrategy}) were applied to predict the normalized \gls{MSE} and \gls{CORR} between the predicted and the actual target values of the \gls{MLP}. Later the performance of the Meta-regressor (\gls{SVM}) was compared with three different benchmarked regression algorithms that were used in the previous work including Linear Regression, \gls{k-NN}\footnote{k=1} and M5 algorithm (\gls{DT} \cite{Quinlan1992}). The experiments revealed that the \gls{SVM} with \gls{RBF} kernel (particularly with $\gamma$=0.1) obtained better performance as a Meta-regressor when compared to the mentioned benchmark algorithms.

\cite{kadlec2009architecture} proposed a generic architecture for the development of on-line evolving predictive systems. The architecture defined an environment that links four classes techniques from the \gls{Machine Learning} area: 1) ensemble methods, 2) local learning, 3) meta-level learning, and 4) adaptability and also the interaction between them. Meta-level learning is discussed in this section whereas adaptability aspects of this paper are discussed in Sections~\ref{sec:adapdation} respectively. 

The Meta-level Learning module of \cite{kadlec2009architecture} architecture was responsible for high-level learning, control and decision making. Meta-level was the most complex but least diverse top layer of the architecture. In this study, a Meta-learner was defined as building a high-level global knowledge of the models which were incrementally grown by applying the evolving architecture to various tasks. The main goal of Meta-level layer was to optimise the predictions in terms of the global performance function which was achieved by 1) controlling the population at lower levels to cover unexplored parts of the input space, 2) looking for relations between algorithm configurations of the paths and the achieved performance, and 3) adapting the combinations in order to reflect the current state of the data. In general, this layer was used to learn the dependency between the pool of learning algorithms and the performance at various levels. Several experiments had been performed using three real-world datasets from the process industry where adaptive and static techniques were compared. The automated data pre-processing and model selection took a lot of the model development effort away from the user. 

An empirical study on rule induction based forecasting method selection for univariate \gls{Time-series} was conducted by \cite{Wang2009}. The study aimed to identify characteristics of a univariate \gls{Time-series} and evaluated the performance of four popular forecasting methods (listed in Table~\ref{table:BaseLearningStrategy}) using a large collection of datasets listed in Table~\ref{table:ExamplesofDatasets}. These two components are integrated into an \gls{Meta-learning} framework which automatically discovers the relations between forecasting methods and data characteristics (shown in Figure~\ref{fig:FigMetaFeatures}). Furthermore, the C4.5 decision tree learning technique was used to generate quantitative rules of \glspl{Meta-features} and categorical rules were constructed using an unsupervised clustering approach.

\cite{LemkeJun2010} investigated applicability of \gls{Meta-learning} for \gls{Time-series} prediction and identified an extensive set of \glspl{Meta-features} that were used to describe the nature of \gls{Time-series}. The feature pool consisted of general statistical, frequency spectrum, autocorrelation, and behaviour of forecasting methods (diversity) measures (see Figure~\ref{fig:FigMetaFeaturesCombination}). These measures were extracted for two sets of datasets from popular TS competitions, see Table~\ref{table:ExamplesofDatasets} for details, and the target was to predict the next 18 observations for NN3\footnote{Neural Network forecasting competition, http://www.neural-forecasting-competition.com}\addtocounter{footnote}{-1}\addtocounter{Hfootnote}{-1} and 56 for NN5\footnotemark. Using these datasets empirical experiments had been performed that had provided the basis for further \gls{Meta-learning} analysis. An extensive list of simple (seasonal), complex (\gls{ARIMA}), structural and computational intelligence (Feed-forward \gls{NN}), and forecast combination methods were used for experimentation which can be seen in Table~\ref{table:BaseLearningStrategy}. From the pool of individual algorithms, \gls{NN} and \gls{MA} performed quite well for NN3 series while for NN5 the \gls{SMAPE}, in general, was quite high where a combination method \emph{variance-based pooling} out-performed all the individual and combination algorithms. At the end three experiments were performed to explore \glspl{Meta-features} using decision trees, comparing various \gls{Meta-learning} approaches (details are given in Table~\ref{table:MetaLearningStrategy}), and simulating NN5 on \emph{zoomed ranking} method and on its combination. This study concluded that the ranking-based combination of forecasting methods outperformed the individual methods in all experiments.

\subsubsection{Clustering}
This section discusses the use of \gls{Meta-learning} in the context of unsupervised learning. 

\cite{DeSouto2008} presented a novel framework that applied an \gls{Meta-learning} approach to clustering algorithms, which was one of the initial efforts towards unsupervised algorithms. The proposed architecture was very similar to the \gls{Meta-learning} approach used to rank regression and classification algorithms. It extracted features of input examples from available datasets and associated them with the performance of the candidate algorithms in clustering that data to construct \gls{Meta-knowledge} database. The \gls{Meta-knowledge} database was used as an input dataset for the Meta-level learning and generated a Meta-model which was used in the selection or ranking of the candidate algorithms at a test mode. Some implementation issues were also addressed which included: 1) the selection of datasets; 2) the selection of candidate clustering algorithms; and 3) the selection of the set of \glspl{Meta-features} that can better represent the problem at the Meta-level. In order to evaluate the framework, a case study using cancer gene expression microarray datasets was conducted. Seven candidate algorithms, listed in Table~\ref{table:MetaLearningStrategy}, and eight descriptive and statistical \glspl{Meta-features} were extracted, namely, \emph{log10} of the number of examples and a ratio of the total number of examples divided by the total number of features, multi-variant normality, a percentage of outliers, a percentage of missing values, the skewness of Hotelling $T^2$-test, a Chip - type of microarray, and a percentage of features that were kept after applying the selection filter. Also, a regression \gls{SVM} algorithm was used as the Meta-learner. The results were compared with the default ranking, where the average performance was suggested for all datasets. The mean and standard deviation of the \gls{Spearman} correlation for both rankings generated by the proposed approach was found to be significantly higher than the default one.

\cite{Soares2009Sep} employed the \cite{DeSouto2008} framework in the ranking task of candidate clustering algorithms in a range of artificial clustering problems with two different sets of \glspl{Meta-features}. The first set had five \glspl{Meta-features} that were calculated using univariate statistics: quartiles, skewness and kurtosis, in order to summarize the multivariate nature of the datasets. This set included \gls{CoV}, \gls{CoV} of second and third quartiles, \gls{CoV} of skewness and kurtosis while the other set had the same first four \glspl{Meta-features} as presented in \cite{DeSouto2008}. In this paper three new candidate clustering algorithms were applied on each learning task that are listed in Table~\ref{table:MetaLearningStrategy} and two Meta-learners were used, i.e., \gls{SVR} and \gls{MLP}. The methodology was evaluated using 160 artificially generated datasets, whose details are discussed in Section~\ref{sec:datasets}. Both Meta-learners were applied to the two sets of \glspl{Meta-features} separately and then compared with the default ranking method. The rankings predicted by the \gls{SVR} and \gls{MLP} methods were found to be significantly higher correlated than the default ranking. However, there was no significant difference between the correlation values of \gls{MLP} and \gls{SVR} methods for both Meta-datasets. Finally, the authors had also highlighted the selection of \glspl{Meta-features} in the context of unsupervised \gls{Meta-learning} as an important issue that could be subjected to further analysis. 

\subsubsection{Discussion and Summary}
There have been several \gls{Meta-learning} systems developed since the beginning of this area. Almost all the systems are developed for algorithm recommendations for the classification and regression tasks. Three main \gls{Meta-features} generation approaches were used in these systems which are listed in Table~\ref{table:ExistingSystems}, where \gls{DSIT} approach is found to be the most widely used. A landmarking based algorithm recommendation system is available as part of the RapidMiner, a commonly used open-source data-mining software. It was part of \gls{PaREn} project and the landmarking functionality is available as an operator in the software. One of the most recent and large-scale projects related to \gls{Meta-learning} was \gls{e-LICO}, the purpose of which was to solve data-mining and data-intensive problems. This project used \gls{Meta-learning} for algorithm recommendation by leveraging the existing systems, i.e., \gls{IDA} and RapidMiner's \gls{DMA} component proposed by \cite{Bernstein2001}. Limitations of those systems are discussed in Table~\ref{table:ExistingSystems}.

Apart from the existing software systems and tools, there have been several studies where \gls{Meta-learning} was used specifically for regression, forecasting, classification or clustering tasks. Several \gls{Meta-features} based problem representations have been proposed for the regression and classification tasks. Most of the comparisons in those studies focused on different \gls{Meta-features} approaches, selection of candidate algorithms, and different sets of Meta-Learners. The problem representation using \glspl{Meta-features} has received the most attention, with landmarking and model-based approaches frequently compared with \gls{DCT} \gls{DSIT} features and outperforming the \gls{DSIT} approach in all reported studies with a significant difference. Not much effort has been dedicated to the model-based approach in the last few years as the landmarking with additional \gls{DSIT} features have been considered as an overall better approach. The landmarking has also been proposed to solve problems other than algorithm recommendations, e.g., \cite{kadlec2009architecture} used a landmarking approach for a recurrent concept extraction. Various studies investigated the applicability of \gls{Meta-learning} for \gls{Time-series} problems including \cite{Prudencio2004}, \cite{Wang2009}, and \cite{LemkeJun2010}. \cite{Prudencio2004} proposed descriptive and statistical features to represent a \gls{Time-series} task to rank various seasonal and \gls{ARIMA} models. Later on \cite{LemkeJun2010} used an extensive list of \gls{Meta-features} covering statistical, frequency spectrum, autocorrelation, and diversity measures for a \gls{Time-series} prediction task. The pool of \gls{Time-series} algorithms contained seasonal, \gls{ARIMA}, structure and computational intelligence, and forecasting combination methods. The features used in this study to represent \gls{Time-series} task at the Meta-level were better as compared to the previous studies.

There have been few studies that applied the \gls{Meta-learning} to clustering algorithms. \cite{DeSouto2008} effort was the initial step in investigating the knowledge representation for unsupervised problems. Landmarking was used to rank several unsupervised candidate algorithms, as listed in Table~\ref{table:MetaLearningStrategy}, combined with eight descriptive and statistical \glspl{Meta-features} which were used to represent unsupervised problems at the Meta-level. Most of them were the same as used in the number of regression and classification problem representations. \cite{Soares2009Sep} employed \cite{DeSouto2008} framework by enhancing the list of landmarkers and proposed two different \gls{Meta-features} representations of an unsupervised task. One of the \glspl{Meta-features} list consisted of features proposed by \cite{DeSouto2008}. The results showed an improvement of the proposed approach over the default base-line, but no significant difference was observed between the two different representations of the unsupervised problems. Finally, the authors had also highlighted the selection of \glspl{Meta-features} in the context of unsupervised \gls{Meta-learning} as an important issue that could be subjected to further analysis. All the existing \gls{Meta-learning} studies discussed in this section have only considered and were applied within stationary environments. Additionally, these systems have the same issue which was discussed in the previous sections that the \gls{Meta-knowledge} dataset did not have a sufficient number of \glspl{Meta-example}. 
\\
\\

\begin{center}
{\small
\begin{longtable}{c|p{3.7cm}|p{5.8cm}|p{3.4cm}}
\caption{Meta-level learning strategy used in various studies}
\label{table:MetaLearningStrategy} \\
\toprule
Research Work & Learning Strategy & Meta-learners & Performance \\
\midrule
\cite{Sohn1999} & \gls{DSIT} approach & Disc, QDisc, LoGID, \gls{k-NN}, Back-propagation, \gls{LVQ}, Kohonen, \gls{RBF}, \gls{INDCART}, \gls{C4.5}, Bayesian Trees & Disc algorithm ranked as top performing algorithm \\

\cite{Lindner1999} & Numeric, Symbolic and Mixed features characterization & \gls{NB}, \gls{MLP}, \gls{RBF}, \gls{CN2}, \gls{ID3}, MC4, T2, Winnow, \gls{OC1}, \gls{OneR}, \gls{Ripper}, \gls{IBL}\footnote{0-4}, \gls{C5.0trees}, \gls{NBT}, \gls{LazyDT}, \gls{PEBLS} & Numeric and mixed features characterization performed better \\

\cite{Bensusan2000b} & Landmarking approach compared with Information-Theoretic characterization & \gls{NB}, \gls{k-NN}\footnote{k=1}, \gls{e-NN}, \gls{DecisionNodes}, Worst Nodes Learner, \gls{RandomlyNodes}, \gls{LDA} & Landmarking (\gls{C5.0rules}) approach outperformed Information-Theoretic \\

\cite{Pfahringer2000} & Landmarking approach compared with \gls{DSIT} characterization & \gls{C5.0trees}, \gls{Ripper}, \gls{Ltree} & Landmarking (\gls{C5.0boost}) performed better than others \\

\cite{Peng2002} & Model-based approach compared with Landmarking and \gls{DSIT} characterization & \gls{k-NN} & Model-based approach outperformed the remaining two \\

\cite{Prudencio2004} & Descriptive and Statistical approach & I: Simple \gls{ES} and Time-delay \gls{NN} and II: \gls{RW}, Holt's linear \gls{ES} (HL), Auto-regressive (AR), NOEMON & I: Simple \gls{ES} and II: NOEMON performed better \\

\cite{DeSouto2008} & Landmarking approach to rank unsupervised learning algorithms & \gls{SL}, \gls{CL}, \gls{AL}, \gls{k-M}, \gls{M}, \gls{SP}, \gls{SNN} & The proposed approach outperformed the default ranking \\

\cite{Guerra2008} & Descriptive and Statistical approach & \gls{SVM} with linear, quadratic, and \gls{RBF} ($\gamma$=0.1, 0.05, 0.01) functions & Normalized \gls{MSE} and \gls{CORR} between predicted and target values \\

\cite{Soares2009Sep} & Landmarking approach to rank unsupervised learning algorithms & \gls{SL}, \gls{CL}, \gls{AL}, \gls{k-M}, \gls{M}, \gls{SNN}, \gls{FF}, \gls{DBS}, \gls{XM} & The proposed approach outperformed the default ranking \\

\cite{Wang2009} & Statistical approach on \gls{Time-series} & \gls{ES}, \gls{ARIMA}, \gls{RW}, \gls{NN} & \\

\cite{LemkeJun2010} & Statistical approach on \gls{Time-series} & \gls{NN}, \gls{DT}, \gls{SVM}, Zoomed ranking (best method and combination) & The proposed approach showed superiority over simple model selection approaches \\

\cite{Abdelmessih2010} & Landmarking approach compared with Descriptive, \gls{DSIT} characterization & \gls{NB}, \gls{k-NN}, \gls{MLP}, \gls{OneR}, \gls{RF} & Landmarking approach (\gls{k-NN}) outperformed others \\

\cite{Rossi2012} & \gls{DSIT} & \gls{RF} & MetaStream outperformed default and ensemble approaches \\

\cite{Rossi2014} & \gls{DSIT} & \gls{RF}, \gls{NB}, \gls{k-NN} & MetaStream outperformed default and ensemble approaches \\
\bottomrule
\end{longtable}}
\end{center}

\subsection{Adaptive Mechanisms} \label{sec:adapdation}
The \gls{Machine Learning} and heuristic search algorithms require tuning of their parameters for a good performance. It can be achieved through off-line sensitivity analysis by testing different parameters to determine their best value in a stationary environment \cite{Sikora2008}. However, the optimal set of values for the parameters keep changing over time in non-stationary environments because of the change in the underlying data distribution where off-line sensitivity analysis becomes ineffective. In a dynamically changing environment domain \gls{Meta-learning} mechanism is considered to be one of the most effective techniques to learn the optimal set of parameters \cite{Sikora2008}. The rest of this section discusses various techniques of acquiring and exploiting \gls{Meta-knowledge} in non-stationary environments, that have been proposed in the context of the existing predictive systems. 

One of the earliest efforts employing an \gls{Meta-learning} based approach to achieve adaptivity in a non-stationary environment was presented by \cite{Widmer1997}. \gls{Meta-learning} was applied in time-varying environments for the purpose of selecting the most appropriate learning algorithm. For a traditional two-level learning model different types of attributes were defined at the Base- and Meta-level. The predictive attributes were used to induce models at the Base-level on raw examples from datasets if there existed a significant correlation between the predictors and the observed class distribution. On the other hand, contextual attributes were employed to identify the current concept associated with the data and systematic changes in their values which indicated a concept drift. These attributes were identified using an \gls{Meta-learning} approach which was proposed in \cite{Widmer1997}. This allowed a learning algorithm to select the examples that had the same context as the training data and newly arrived examples. These conceptual clues helped in adapting the systems faster by filtering the historical instances used for training that had the same context as the newly arrived instances. The proposed technique was evaluated by comparing two operational systems at the Meta-level that differed in the underlying learning algorithm as well as their way of processing contextual information including METAL(B) that used a Bayesian classifier and METAL(IB) that was based on instance-based learning. The instance-based learner was used in four variants which included: 1) context-relevant instance selection; 2) instance weighting; 3) feature weighting; and 4) combination of instance and feature weighting. The general conclusion of numerous experiments that were performed using real-world and synthetic datasets was that \gls{Meta-learning} produced quite a significant improvement over the existing approaches for changing environments. Additionally, from the results, it could be observed that the METAL(B) approach proved to be effective in domains (datasets) with high noise rates and several irrelevant attributes whereas the instance-based approach showed higher accuracy for the remaining domains.

\cite{Klinkenberg2005} proposed an \gls{Meta-learning} framework for automatically selecting the most promising algorithm and its parametrization at each step in time where the data was arriving in batches. For each batch a set of \glspl{Meta-features} (as listed in Table~\ref{table:AdaptationTech}) were extracted directly from the raw data which was used in the \gls{Base-learning} to create a Meta-example. A number of Meta-examples were used to induce a Meta-learner whenever a new batch became available, which in turn, helped in predicting the best learning algorithm and the best set of instances at a given time point. The \glspl{Meta-features} used in this work were more relevant to the problem under analysis. Furthermore, this work also investigated the aspects used to speed-up the algorithm selection process using the proposed \gls{Meta-learning} approach without losing the gained reduction in the error rate. The proposed drifting concept approaches, i.e., adaptive time window and batch selection strategy, were evaluated by comparing them with three non-adaptive mechanisms: 1) full memory, 2) no memory, and 3) fixed-size window. The experiments were performed using two real-world problems: 1) information filtering of unstructured business news data, and 2) predicting the business cycle from the economics domain. For the business news dataset, both adaptive techniques outperformed trivial non-adaptive approaches. Two evaluations were performed for the business cycle dataset where the data was split into 5 and 15 equally sized batches where the fixed size window approach performed slightly better than the adaptive techniques. 

\cite{Sikora2008} proposed an \gls{Meta-learning} mechanism to learn the optimal parameters while the learning algorithm was trying to learn its target concept in a non-stationary environment. \gls{Meta-learning} was used to tune a temperature ($\tau$) parameter of the Softmax \gls{RL} algorithm using a Boltzmann distribution. Moreover, the time-weighted method had been used where the action value estimates were the sample average of prior rewards. The Softmax algorithm became a random search for a higher $\tau$ value, whereas for a low value it approached a greedy search. The effectiveness of the proposed \gls{Meta-learning} algorithm was evaluated by dynamically learning the optimal value of $\tau$ using two case studies: 1) k-Armed bandit - the classic \gls{RL} problem, and 2) bidding strategy - stylized e-procurement problem. In the k-Armed bandit problem the variable \emph{k} was defined as actions available to an agent and each action returned a reward from a different distribution. In this work (\emph{k}=) 10 actions (1,...,10) were available to an agent where each action returned a reward using Normal distribution. The effectiveness of \gls{Meta-learning} in a non-stationary environment was tested by rotating the reward distributions among the 10 actions. The algorithm was tested with three different temperature parameter values of 5, 50 and 500 for both stationary and dynamic environments. For the stationary environment, the performance of $\tau$=5 approached the best action with a maximum average reward. As the environment became more and more dynamic these awards kept falling. In contrast, the performance of the \gls{Meta-learning} algorithm returned better rewards in both environments as well as responded faster to the changes in the environment. The bidding problem was analysed as a 2 player symmetric game (2 homogeneous sellers) with \emph{n} actions, where \emph{n} was the variable cost (price) range split into equally sized bands. One of the sellers was modelled using the Softmax \gls{RL} algorithm while the other one was supposed to be using different learning algorithms, i.e., $\epsilon$-greedy - a genetic algorithm proposed by \cite{Goldberg1989}. The same three values of $\tau$ were used for both stationary and dynamic environments, where the stationary environment produced the best result for the lowest value of temperature. However, no single value of temperature did best in the dynamic environment, while \gls{Meta-learning} algorithm approached the best reward for both environments. Furthermore, it was observed from the experiments that the best value of $\tau$ was achieved from \gls{Meta-learning} approach in all the scenarios. 

\cite{kadlec2009architecture} architecture supports life-long learning by providing several adaptation mechanisms across computational path level (preprocessing methods followed by individual base-level algorithms), path combination level (a combination of base-level algorithms) and a Meta-level hierarchical structure. There were four adaptation loops defined across various levels of hierarchy including the self-adaptation capability of the computational and combination layer, whereas the remaining two loops connected the Meta-level layer to the lower layers. These feedback loops helped the proposed architecture to keep the validity of the models in changing environments. It could be achieved by switching particular modules to the incremental mode. The computational path level adaptation loop consisted of the predictions feedback which were compared to the actual (target) values. Whereas at the path combination level the combinations were represented in the same way as in the computational path, which was a benefit of this representation that and meant that similar adaptation mechanisms could be applied at different levels. In the case of weighted combinations, the contribution of particular computation paths was dynamically changed to the final prediction by modifying the weights. A Meta-level adaptation influenced the dynamic behaviour of the entire architecture. At this level, the performance measures were gathered from all levels of the architecture together with the global performance. It allowed us to analyse the performance achieved across various levels and also to estimate the influence of the changes at different states of the model. Several experiments demonstrated that the variety of adaptation mechanisms applied at different levels may have a significant effect on the performance of the models. One of the key contributions of the proposed architecture was the opening of a large space for future research that could focus on the interaction between different techniques, dynamic behaviour, implementation of novel adaptation techniques and meta-level methods.

\begin{figure}[ht!] 
    \centering
	\includegraphics[viewport=4 225 650 790, clip, scale = 0.815]{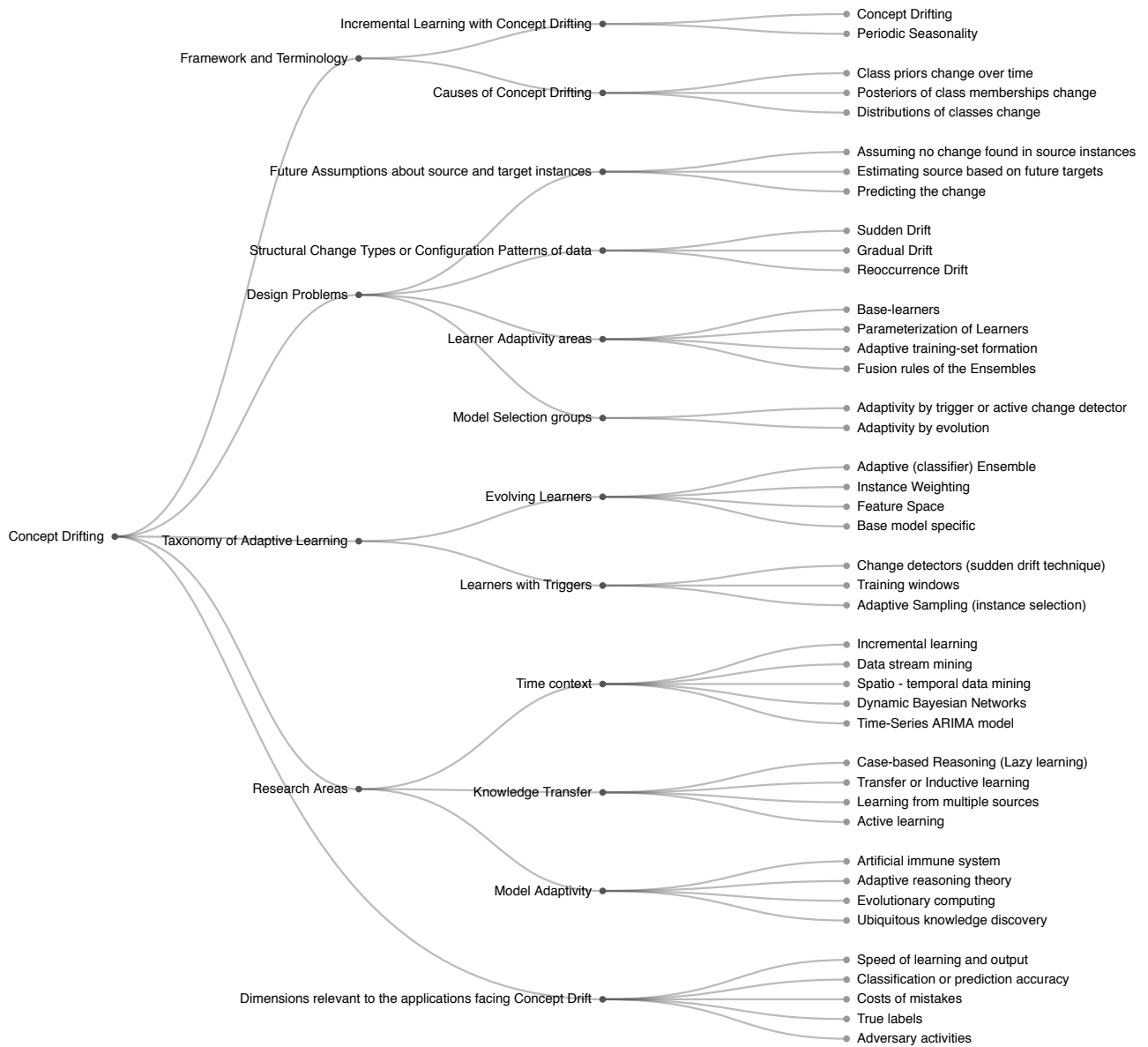} 
	\caption{Learning under Concept Drifting \cite{Zliobaite2010}}
	\label{fig:FigConceptDrifting}
\end{figure}

A comprehensive framework, design problems, the taxonomy of adaptive learning, and different areas of learning under concept drift were presented by \cite{Zliobaite2010}. The proposed framework was used to analyse the problem of the training set formation where two areas, i.e., 1) incremental learning; and 2) causes of concept drift were discussed. The incremental learning explained the difference between concept drift and periodic seasonality with examples while the causes of concept drift were elaborated on using Bayesian decision theory, where three causes were highlighted that might change over time. There were four design sub-problems and techniques addressed within the framework that needed to be solved: 1) future assumptions about source and target instances; 2) structural change types or configuration patterns of data over time; 3) identified four key learner adaptivity areas, and 4) model selection which was further categorized into two different groups. The taxonomy of concept drift learners was categorized as an evolving learner where four methods were proposed and the methods that determined how the models or instances were to be changed at a given time were grouped separately under a triggering concept. In the end, three major research areas were outlined: 1) time context; 2) transfer learning by gaining knowledge from a similar type of past problems; and 3) models that have properties of adaptation incorporated into learners. Also, several dimensions that are relevant to the applications implementing concept drift were defined. Figure~\ref{fig:FigConceptDrifting} presents all the key areas and available solutions of learning under concept drift. 

An \gls{Meta-learning} approach for periodic and automatic algorithm selection for time-changing data, named Meta-Stream, was presented by \cite{Rossi2012}. A Meta-classifier was periodically applied to predict the best learning algorithm for a new unlabelled chunk of data. General DSIT \glspl{Meta-features} of the Travel Time Prediction (TTP) problem were extracted from the historical and new data (see Figure~\ref{fig:FigMetaFeatures}) and mapped together with their predictive performance computed from different models to induce the Meta-classifier. Experiments were performed to compare the performance of the MetaStream to the default trial-and-error approach for both static and dynamically updating strategies at the Meta- and Base-levels. Moreover, the Base-level MetaStream and Default results were compared with the dynamic Ensemble approach. The learning strategy adopted at the Base-level can be seen in Table~\ref{table:BaseLearningStrategy}, also the training window ($\omega$) of 1000 instances with a step size ($\lambda$) of 1 was used at this level. The Meta-level learning strategy was presented in Table~\ref{table:MetaLearningStrategy}. The \glspl{Meta-example} labelled as tie were investigated separately by keeping and discarding them from the training and test sets. The empirical results showed that the MetaStream outperformed the baseline and ensemble approaches with a significant margin in most of the cases for both stationary and dynamic environments. In general, the two pairs of algorithms, e.g., \gls{RF}-\gls{CART} and \gls{SVM}-\gls{CART} were found to be the best algorithms for TTP problem. Finally, the authors also realized that the \glspl{Meta-features} should be related to the non-stationary data problem rather than characteristics that were extracted for the traditional \gls{Meta-learning} problems.

\cite{Rossi2014} extended their original work \cite{Rossi2012} in two main directions: 1) instead of selecting only a single algorithm, a combination of multiple regressors could be selected when the average of the predictions performed better than the individual; and 2) more comprehensive experimental evaluation was performed by adding another real-world problem -  \emph{Electricity Demand Prediction (EDP)} (see Table~\ref{table:ExamplesofDatasets}). Furthermore, the list of \glspl{Meta-features} extracted from the data was also enhanced in this work, as listed in Table~\ref{table:MetaStream}. The characteristics were extracted separately from the training and evaluation windows because the training window had target information available from where supervised characteristics could be extracted, i.e., information about the relationship between the predictive and target variables. The pool of Base- and Meta-level algorithms with their configurations are listed in Table~\ref{table:BaseLearningStrategy} and Table~\ref{table:MetaLearningStrategy} respectively. The experimental results showed that for the TTP dataset the pair of regressors, regardless of the presence of the tie resolution strategy, outperformed the default and ensemble-based approaches. However, in the case of EDP, the MetaStream clearly outperformed the default but was worse than the ensemble which could lead to a conclusion that the observations made for pairs of regressors were also valid for multi-regressors. Moreover, a slightly higher error rate was recorded for \gls{RF} Meta-learner of the MetaStream than the default but was lower than the ensemble approach for the TTP dataset, whereas for the EDP dataset the MetaStream outperformed the default but was worse than the ensemble. These results showed that the MetaStream was able to select the best algorithm more accurately than the baseline trial-and-error and ensemble-based approaches in a time-changing environment.

\begin{center}
{\small
\begin{longtable}{l|c|c}
\caption{Meta-features used in MetaStream to characterize the data}
\label{table:MetaStream} \\
\toprule
Meta-features & Training window & Selection window \\
\midrule
Average, Variance, Minimum, Maximum and Median of continuous features & \tick & \tick \\

Average, Variance, Minimum, Maximum and Median of the target & \tick & - \\

Correlation between numeric features & - & \tick \\

Correlation of numeric attributes to the target  & \tick & - \\

Possibility of existence of outliers in numeric features & - & \tick \\

Possibility of existence of outliers in the target  & \tick & - \\

Dispersion gain & \tick & - \\

Skewness of numeric features & - & \tick \\

Kurtosis of numeric features & - & \tick \\
\bottomrule
\end{longtable}}
\end{center}

\subsubsection{Discussion and Summary} 
This section covered the adaptability mechanisms of a number of existing systems using MLL approaches. In these studies, the main focus was put on the applicability of \gls{Meta-learning} particularly in the context of non-stationary environments. \gls{Meta-learning} can be beneficial in such a case by minimizing the processing time that is consumed to periodically train the model, extracting recurring concepts, automatically detecting concept drift and estimating dynamic adaptive window size, which in turn can generate accurate predictions in dynamic environments. However, applying \gls{Meta-learning} to support an adaptive mechanism is a recent and emerging area. As a result, most of the research works use the same \glspl{Meta-features} for a time-varying environment as for the stationary environments. If \gls{Meta-learning} is introduced in a system then the overall performance of such a system becomes dependent on an appropriate representation of the problem at the Meta-level in the form of extracted, informative \glspl{Meta-features}. The drawback of using a set of \glspl{Meta-features} which are usually used in a stationary environment is that the entire target dataset should be available at once when \gls{Meta-learning} is applied to find the best algorithm for that dataset. This is not normally the case for streaming data and the unavailability of target variables makes the calculation of some useful \glspl{Meta-features} impossible.

\cite{Widmer1997}'s work on applying \gls{Meta-learning} for non-stationary environments is considered to be the earliest effort. It addressed two key areas in the context of dynamic environments: 1) dynamic tracking of changes; and 2) extraction of recurring concepts. The problem representation in \cite{Widmer1997} was quite general as very few predictive and contextual \glspl{Meta-features} were extracted. However, neither of the two proposed \gls{Meta-learning} approaches performed better than the default for several domains. \cite{Klinkenberg2005} used different \gls{Base-learning} algorithms which were automatically selected at the Meta-level. Additionally, the Meta-level approach for adaptive time window and recurring concept extraction for the target concept was part of the research. The research was one of the initial efforts to represent an adaptivity problem with the relevant \glspl{Meta-features} rather than using general features that were usually productive for the stationary environment. Although these features (as listed in Table~\ref{table:AdaptationTech}) were not sufficiently expressive to represent a non-stationary environment at the Meta-level, they were still better than general features (used to represent stationary problems) as evidenced by the experiments which showed a significant improvement. 

\cite{Sikora2008} proposed a reinforcement learning approach to address the automatic algorithm recommendation problem using \gls{Meta-learning} in a non-stationary environment. The focus of the research was to find the optimal value of the Softmax algorithm's parameter $\tau$ where it would recommend the best algorithm for the target concept at the Meta-level. The same deficiency was observed in this work that the non-stationary problem representation was not addressed in sufficient detail and focus was only on the algorithm recommendation using \glspl{Meta-features} which were proposed for static data. \cite{kadlec2009architecture} proposed a life-long learning architecture that provided several adaptation mechanisms across a pool of candidate learning algorithms and their combinations. The dynamic behavior of the entire architecture was analysed at the Meta-level where the global performances as well as information from both pools could be analysed to estimate the influence of the changes at different levels of the model. The decrease in the prediction ability of a local model below a certain level was considered as a new concept which led to building a new receptive field. The landmarking approach was quite simple and effective to detect concept drift, and based on that, periodically train a new local predictor. The effectiveness of \gls{Meta-learning} for the two mentioned areas was supported by improved results recorded from two case-studies. 

\cite{Rossi2012} approach was quite similar to \cite{Klinkenberg2005} where periodic algorithm selection for a time-changing data was proposed. Similarly to various other studies, the authors computed the \gls{DSIT} \glspl{Meta-features}. Even though the Meta-level approach performed better than the Base-level, there was no comparison shown with the other \gls{Meta-learning} systems from where it could be concluded that even the general representation of the problem could work for a non-stationary environment. The problem representation using general \glspl{Meta-features} was a drawback of this effort which was subsequently attempted to rectify in \cite{Rossi2014}. The authors computed separate \glspl{Meta-features} for historical and incoming data. As the target variable was not available in the incoming data the unsupervised features were computed for the data available in the evaluation window. The performance of the proposed approach was better than the \gls{Base-learning} and worse than an ensemble-based approach but despite this, it was considered to be a good effort towards representing a time-varying problem at the Meta-level. In almost all the studies that are discussed in this section \gls{Meta-learning} outperformed the \gls{Base-learning} methods. However, a common drawback has been observed in the problem representation area at the Meta-level for time-varying data. Most of the work used general \glspl{Meta-features} whereas only some tried to focus on this area by proposing some features for the non-stationary data.

\begin{center}
{\small
\begin{longtable}{c|p{4.5cm}|p{7.5cm}}
\caption{Adaptive mechanisms used in previous studies}
\label{table:AdaptationTech} \\
\toprule
Research Work & Adaptivity mechanisms addressed & Meta-features/Parameters \\
\midrule
\cite{Widmer1997} & Recurring concept extraction & window size=100 and significance level=0.01 \\

\cite{Klinkenberg2005} & Recurring concept extraction, adaptive time window, periodic algorithm selection & \pbox{20cm}{No. of batches used for training at the previous batch \\ No. of non-interrupted most recent training batches \\ Most successful learner on the previous batch \\ Most successful learner overall on all batches seen so far} \\

\cite{kadlec2009architecture} & Concept drift detection and Periodic algorithm selection & Landmarking \\

\cite{Rossi2012} & Periodic algorithm selection & \pbox{20cm}{\gls{Machine Learning}: $\omega$=1000, $\lambda$=1, $\eta$=0 \\ \gls{Meta-learning}: $\omega$=300, $\gamma$=25, $\lambda$=1, $\eta$= 0} \\

\cite{Rossi2014} & Periodic algorithm selection (with more relevant representation of the non-stationary problem) & \pbox{20cm}{TTP dataset: \pbox{20cm}{\gls{Machine Learning}: $\omega$=1000, $\lambda$=1, $\eta$=2 \\ \gls{Meta-learning}: $\omega$=300, $\gamma$=24, $\lambda$=1, $\eta$=0} \\ EDP dataset: \pbox{20cm}{\gls{Machine Learning}: $\omega$=672, $\lambda$=336, $\eta$=0 \\ \gls{Meta-learning}: $\omega$=300, $\gamma$=25, $\lambda$=1, $\eta$=0}} \\
\bottomrule
\end{longtable}}
\end{center}

\section{Research Challenges} \label{sec:challenges}
The goal of \gls{Meta-learning} is to analyse and recommend the best methods and techniques for a problem on the basis of previously solved problems and without or with minimal intervention of human experts \cite{Duch2011}. The existing approach of analysing the problem and selecting the best learning algorithm is to apply a wide range of algorithms, with many possible parametrizations, on a problem simultaneously and then select an algorithm from a ranked list based on performance estimates like accuracy, execution-time, etc. Also choosing the best algorithm for a specific problem in an ever increasing number of models and their almost infinite configurations is a challenging task. Even with sophisticated and parallel learning algorithms, the computational power in terms of the execution-time, memory, and the overall human effort are still one of the biggest limitations. Every task leads to new challenges and demands dedicated effort for detailed analysis and modelling. 

The main theme of this work is research on \gls{Meta-learning} strategies and approaches in the context of adaptive multi-level, multi-component predictive systems for time-varying environments. In these systems, there are multiple areas where \gls{Meta-learning} can be used to efficiently recommend the most appropriate methods and techniques. Therefore three areas of evolving predictive systems dealing with streaming data have been identified where the applicability of \gls{Meta-learning} can be an effective and efficient approach. These are listed below:

\begin{enumerate}[1.]
    \item Pre-processing Steps Recommendation: \\
    \gls{Meta-learning} can be applied to find the most appropriate combination of pre-processing steps for \gls{Meta-knowledge} dataset. As \gls{Meta-learning} is proposed for four different areas within a system which means in case a concept drift is detected a maximum of four \gls{Meta-knowledge} datasets, which will be representing different problems, will require pre-processing. The applicability of \gls{Meta-learning} on changing environment requires dynamically growing \gls{Meta-knowledge} dataset where a fixed set of pre-processing methods and techniques can be ineffective. Alternatively, trying various pre-processing methods and techniques to find the best combination for the current concept will make the entire system ineffective. Instead of spending time on testing various methods on every concept drift detection \gls{Meta-learning} helps to instantly and optimally recommend the best pre-processing steps for the current concept. 
    
    \item Algorithm Recommendation: \\
    Finding the optimal algorithm for a dataset is a traditional application of \gls{Meta-learning} [Giraud-Carrier, 2008]. Automatic discovery of optimal algorithm can be beneficial for both stationary and particularly non-stationary environments where it can help in minimizing the processing time which is usually spent on the rigorous testing of various learning algorithms with their different parametrizations. \gls{Meta-learning} can recommend the optimal learning algorithm and parametrization instantly.
    
    \item Recurring Concepts Extraction: \\
    In a non-stationary environment, the underlying distribution of the incoming data keeps changing which makes the most recent historical data ineffective to retraining the model for the batch of data available in the evaluation window. Using \gls{Meta-learning} the historical batches (concepts) of data can be extracted from \gls{Meta-knowledge} dataset which can be used as effective data for training of the current concept. This process can be named as \emph{Reverse Knowledge Extraction} where \glspl{Meta-features} of the current concept can be used to extract the \glspl{Meta-example} of relevant concepts from \gls{Meta-knowledge} datasets. These \glspl{Meta-example} will ultimately lead to extract the model that can be the best representation of the current concept. This model can be retrained to incorporate a new concept in the existing model.
    
    \item Dynamic Concept Drifting and Adaptivity Mechanism Parameters: \\
    The most process and memory-intensive task in the system is model training which has to be performed on the identification of every new concept. In an adaptive mechanism retraining of model is usually triggered by a change detection process where intelligent triggering can maximize the overall system efficiency. \gls{Meta-learning} can help to automatically detect the concept drift and trigger the algorithm retraining process instantly. For instance, the \glspl{Meta-features} of incoming data can be computed as well as cumulated on the arrival of every batch and simultaneously compared with the set of \glspl{Meta-example}, from \gls{Meta-knowledge} dataset, whose learning algorithm (used as target variable in \gls{Meta-knowledge}) is used to score the current batches of data. The concept drift is detected at Meta-level if the \gls{Meta-example} of the current concept does not match with the cluster of \glspl{Meta-example} whose learning algorithm is currently selected.
    
    Using the same technique the dynamic adaptive mechanism parameters problem within the non-stationary environment can also be addressed. The static parameters of adaptive mechanism, i.e., training window size, evaluation window size, step size, and delay, would be ineffective for the dynamic environments where the underlying distribution of incoming data keeps changing. A \gls{Meta-knowledge} dataset can be gathered containing the various parameters of the adaptive mechanism as \glspl{Meta-features} and mapped with the algorithm or combination that is performing the best for those parameters. Based on the currently selected algorithm the appropriate set of parameters can be extracted from the \gls{Meta-knowledge} dataset. This task is can be named as \emph{Reverse Meta-level Learning}.
\end{enumerate}

The first potential area where \gls{Meta-learning} can be leveraged to find the most appropriate combination of pre-processing steps is already under investigation within the INFER project. So this area is excluded from the scope of this research. The applicability of \gls{Meta-learning} on the remaining three proposed areas leads to several research questions which are listed below.

\begin{enumerate}[1.]
\item Gathering examples of datasets for building Meta-knowledge database

\begin{enumerate}[i.]
    \item The time-changing environments require dynamic \gls{Meta-knowledge} databases which must be updated with the \glspl{Meta-features} of different batches of data having different distribution. A dynamic \gls{Meta-knowledge} database keeps on growing with the \gls{Meta-example} of new concepts. Apart from the dynamically growing database, which will be empty in the initial phase of the system and will gradually build-up, is there need of static \gls{Meta-knowledge} database, atleast for this phase, which is usually used by traditional \gls{Meta-learning} systems?
    
    \item In absence of static \gls{Meta-knowledge} database, \gls{Meta-learning} would be ineffective specifically in the initial phase of the system until a sufficient amount of concept drifts are identified. At Meta-level this impact would greatly effect because one \gls{Meta-example} in \gls{Meta-knowledge} dataset will be extracted from one concept drift which might consist of several batches of data. What could be the alternative of static \gls{Meta-knowledge} database so that the system can leverage from \gls{Meta-learning} process even in the initial stage of the system?
    
    \item The static \gls{Meta-knowledge} database could be a potential solution of the above challenge, but it raises another research challenge that what sources and techniques will be used to gather examples of datasets, e.g., is it possible to find enough real-world problems to extract sufficient \glspl{Meta-example} for \gls{Meta-knowledge} database or synthetic data will be used to produce more \glspl{Meta-example}?
    
    \item If real-world datasets, which are quite rare and hard to find, would be insufficient then generating synthetic data could be a potential solution. In that case what type of techniques will be used to generate examples of synthetic datasets or else by transforming the existing \glspl{Meta-example}, which are generated by real-world datasets, the \gls{Meta-knowledge} database will be enhanced?
\end{enumerate}

\item Base-level learning to compute performance measures for Meta-examples

\begin{enumerate}[i.]
    \item Base-level Learning is used to build predictive models using examples of datasets to compute a comprehensive set of performance measures. What type of strategy will be used to select the best learning algorithm and its parametrization, i.e., selection, ranking, combination?
    \item To what level of granularity the learning algorithm parametrization would be sufficient enough for an effective \gls{Meta-learning} process, e.g., how to deal with continuous parameters, numerous parameters for a learner? 
    \item What performance measures will be used to rank different algorithms for a dataset, i.e., accuracy, run-time speed?
\end{enumerate}

\item Feature generation and selection to represent a problem at Meta-level

\begin{enumerate}[i.]
    \item Would the traditional \gls{Meta-features} generation approach be the better representation of the three different problems where \gls{Meta-learning} would be able to outperform Base-learning?
    \item From the traditional \gls{Meta-features} generation approaches what techniques can be used to represent the problem of the mentioned areas?
    \item The traditional \gls{Meta-features} generation approaches, which are only specialized for algorithm recommendation task, would be adequate to represent three new areas of the system or based on the complexity of a problem a new problem representation would be required?
    \item Within a \gls{Meta-features} generation approach what set of \glspl{Meta-features} could be significant to better represent a problem? What statistical methods would be used to dynamically select significant \glspl{Meta-features} for a batch of data?
    \item In a non-stationary environment, the target variable would not be available at the time of algorithm selection at Meta-level. It will restrict computing a few important \glspl{Meta-features}, e.g., the correlation between target and predictors. What would be the approach of selecting a significant set of \glspl{Meta-features} in the absence of a target variable? 
    \item In a later stage, when the target variable becomes available then how \gls{Meta-knowledge} database will be updated, i.e., retraining with new \glspl{Meta-features}, where the target variable will be involved?
\end{enumerate}

\item Representation and storage of dynamically growing complex Meta-Knowledge database

\begin{enumerate}[i.]
    \item A single \gls{Meta-knowledge} database consisting of numerous \glspl{Meta-features} would be productive to represent all three areas or separate \gls{Meta-knowledge} databases, specialized to solve a specific problem, would be gathered at different levels and at different times?
    \item What level of granularity would be required for the better representation of a problem? For instance, the target variable of the \glspl{Meta-example} would be only the best learning algorithm, all the available algorithms with their rankings, algorithm parametrization.
    \item What type of performance measures will be stored in \gls{Meta-knowledge} database for three different areas, e.g., accuracy, run-time speed? For instance, the run-time speed measure might be useful particularly for a non-stationary environment which helps to identify accurate as well as an efficient learning algorithm.
\end{enumerate}

\item Meta-level Learning

\begin{enumerate}[i.]
    \item What type of different learning strategies and algorithms would be used at Meta-level to efficiently search the best algorithm from \gls{Meta-knowledge} database? 
    \item If \gls{Meta-learning} process recommends an entirely new algorithm for a new concept then what would be the impact of replacing the current algorithm instantly? 
    \item Replacing the algorithm for a concept will enhance the overall performance of the system in all the cases or is there a possibility that replacing algorithms may disturb the accuracy of the system?
    \item \gls{Meta-learning} is proposed for three most important areas within the system, would it be effective enough to rely a lot on Meta-level learning?
\end{enumerate}
\end{enumerate}

\section{Summary} \label{sec:summary}
This literature review and identification of key research challenges have been focused on the detailed study of existing \gls{Meta-learning} concepts and systems for both stationary and non-stationary environments. We are particularly interested in fully automating the process of building, deployment and maintenance of potentially complex multi-component, multi-level evolving predictive systems operating in continuously changing environments, as described in some of our previous publications and those resulting from the INFER project.

The review of the existing research has been structured into the coverage of five key components of an \gls{Meta-learning} system: (i) Available real and synthetic datasets for modelling at the Meta-level; (ii) Meta-features generation and selection approaches; (iii) Base-level learners as an input to the Meta-learning; (iv) Meta-learning; (v) Meta-learning based adaptive mechanisms for non-stationary environments. 

There are various methods to gather \gls{EoD} discussed though all of them have some limitations. Similarly, several Meta-feature generation techniques are reviewed from previous work though the majority of them have been introduced in the context of and are suitable for a stationary \gls{Meta-learning} system. Hence the applicability and effectiveness of such Meta-features for non-stationary environments remain an open research question. A consistently and systematically evaluated performance of base-models on \glspl{EoD} forms a critical part of a reliable input data (i.e. label or target variable) for the \gls{Meta-learning}. Collecting such performance data is the most time and processor-intensive task especially if numerous configurations and parametrisations of base-learners are to be adequately taken into account. Such a reliable collection of previously solved problems with thorough benchmarking of base-learners suitable for \gls{Meta-learning} does not currently exist and remains an open challenge.

A number of previously proposed \gls{Meta-learning} systems have been discussed in detail which included the application of \gls{Meta-learning} to both supervised and unsupervised learning problems. The development and evolution of the \gls{Meta-learning} field in the last three decades has been discussed and various systems have been compared with the previous ones. However, there are very few systems that have been targeted towards and can deal with non-stationary problems which are our main areas of interest. It is only in the last five years that non-stationary \gls{Meta-learning} have been receiving some interest. The primary focus has been on the problem representation of streaming data at the Meta-level.

There are multiple roles for Meta-learning in the scope of the INFER project and the developed automated and autonomous predictive modelling system and approaches working in continuously changing environments which we are intending to explore in our continuing research in this area.

\newpage
\appendix
\section{Meta-features} \label{sec:appendix1}

\begin{center}
{\tiny
\begin{longtable}{p{4.16cm}|p{0.25cm}|p{0.25cm}|p{0.25cm}|p{0.25cm}|p{0.25cm}|p{0.25cm}|p{0.25cm}|p{0.25cm}|p{0.25cm}|p{0.25cm}|p{0.25cm}|p{0.25cm}|p{0.25cm}|p{0.25cm}|p{0.25cm}|p{0.25cm}|p{0.25cm}|p{0.25cm}}
\caption{Meta-features used in various studies}
\label{table:MetaFeaturesTable} \\
\toprule
Meta-Features & \cite{Rendell1987}, \cite{Rendell1990} & \cite{King1995} & \cite{Sohn1999} & \cite{Lindner1999}, \cite{Berrer2000}, \cite{Giraud-Carrier2005} & \cite{Bensusan2000a} & \cite{Bensusan2000b} & \cite{Pfahringer2000} & \cite{Todorovski2002} & \cite{Peng2002} & \cite{Kopf2002} & \cite{Brazdil2003} & \cite{Prudencio2004} & \cite{Prudencio2008}, \cite{Guerra2008} & \cite{Wang2009} & \cite{LemkeJun2010} & \cite{Abdelmessih2010} & \cite{Rossi2012} & \cite{Feurer2014} \\
\midrule
\multicolumn{19}{l}{Descriptive Meta-features} \\ \hline
\gls{Classes} & & \tick & \tick & \tick & & & \tick & & & & & & & & & & & \tick  \\
Frequency of most common class & \tick & & & & & & \tick & & & & & & & & & & &  \\
\gls{Features} & \tick\footnote{only these two features are used in \cite{Rendell1987}, they are also part of \cite{Rendell1990}}\addtocounter{footnote}{-1}\addtocounter{Hfootnote}{-1} & \tick & \tick & & & & & & & & & & & & & & & \tick  \\
\gls{Instances} & & \tick & \tick & \tick & & & \tick & & & & \tick & \tick & & & \tick & & & \tick  \\
Dataset Dimentionality & & & & & & & & & & & & & & & & & & \tick \\
\gls{TrainingInstances} & \tick\footnotemark & \tick & \tick & & & & & & & & & & \tick\footnote{Log}\addtocounter{footnote}{-1}\addtocounter{Hfootnote}{-1} & & & & &  \\
\gls{TestInstances} & \tick & \tick & \tick & & & & & & & & & & & & & & &  \\
Sampling Distribution & \tick & \tick & & & & & & & & & & & & & & & &  \\
\gls{BinaryFeatures} & \tick & \tick & \tick & & & & & & & & & & & & & & &  \\
\gls{NumericFeatures} & \tick & & & \tick & & & \tick & & & & & & & & & & & \tick  \\
\gls{NominalFeatures} & \tick & & & \tick & & & \tick & & & & & & & & & & & \tick \\
Proportion of binary features (\gls{BinaryFeatures}/\gls{Features}) & & & \tick & & & & & & & & & & & & & & &  \\
Proportion of nominal features (\gls{NominalFeatures}/\gls{Features}) & & & & \tick & & & & & & & \tick & & & & & & & \tick  \\
Span of nominal values & & & & \tick & & & & & & & & & & & & & &  \\
Average of nominal values & & & & \tick & & & & & & & & & & & & & & \tick  \\
Training instances to features ratio (\gls{Instances}/\gls{Features}) & & & \tick & & & & & & & & & & \tick\footnotemark & & & & &  \\
Proportion of training instances (\gls{TrainingInstances}/\gls{Instances}) & & & \tick & & & & & & & & & & & & & & &  \\ \hline

\multicolumn{19}{l}{Statistical Meta-features}  \\ \hline
Relative probability of missing values & & & & \tick & & & & & & & \tick & & & & & & & \tick  \\
Instances with missing values & & & & \tick & & & & & & & & & & & & & & \tick  \\
Proportion of features with outliers & & & & \tick & & & & & & & \tick & & & & & & \tick &  \\
Mean \gls{SKEW} & & \tick & \tick & \tick & & & & & & & & & & \tick & \tick\footnote{of the series}\addtocounter{footnote}{-1}\addtocounter{Hfootnote}{-1} & & \tick & \\
Mean \gls{KURT} & & \tick & \tick & \tick & & & & & & & & & & \tick & \tick\footnotemark & & \tick & \\
Average & & & & & & & & & & & & & & & & & \tick &  \\
Variance & & & & & & & & & & & & & & & & & \tick &  \\
Minimum & & & & & & & & & & & & & & & & & \tick &  \\
Maximum & & & & & & & & & & & & & & & & & \tick &  \\
Median & & & & & & & & & & & & & & & & & \tick &  \\
Correlation between predictor and target & & & & & & & & & & & & & & & & & \tick &  \\
\gls{StdDev} of the class distribution & & & & \tick & & & & & & & & & & & \tick\footnote{of the de-trended series} & & & \tick \\
\gls{SDRatio} & & \tick & \tick & \tick & & & & & & & & & & & & & &  \\
\gls{CANCOR} & & \tick & \tick & \tick & & & & & & & \tick & & & & & & & \\
\gls{DiscFunc} & & & & \tick & & & & & & & & & & & & & &  \\
\gls{CORR} & & \tick & \tick & & & & & & & & & & & & & & &  \\
\gls{FRACT} & & \tick & \tick & \tick & & & & & & & & & & & & & &  \\
\gls{Wlambda} & & & & \tick & & & & & & & & & & & & & &  \\
Default Accuracy & & & & \tick & & & & & & & & & & & & & &  \\
coefficient of variation (COEF-VAR) & & & & & & & & & & & & \tick & & & & & &  \\
absolute value of the \gls{SKEW} and \gls{KURT} coefficient & & & & & & & & & & & & \tick & & & & & & \tick  \\
\gls{Time-series} mean absolute values of first 5 auto-correlations (Mean-CORR)& & & & & & & & & & & & \tick & & & & & &  \\
\gls{Time-series} test of significant auto-correlations (TAC) & & & & & & & & & & & & \tick & & & & & &  \\
\gls{Time-series} significance of the 1, 2, and 3 Auto-correlation (TAC-1,2,3) & & & & & & & & & & & & \tick & & & & & &  \\
\gls{Time-series} test of Turning Points for randomness & & & & & & & & & & & & \tick & & & & & &  \\
\gls{Time-series} first coefficient of auto-correlation (AC1) & & & & & & & & & & & & \tick & & & & & &  \\
\gls{Time-series} type & & & & & & & & & & & & \tick & & & & & &  \\
\gls{Time-series} trend & & & & & & & & & & & & \tick & & \tick & \tick\footnote{\gls{StdDev} of the series / \gls{StdDev} of the de-trended series} & & & \\
\gls{Time-series} turning point & & & & & & & & & & & & \tick\footnote{ratio} & & & \tick & & &  \\
\gls{Time-series} \gls{DW} & & & & & & & & & & & & & & & \tick & & &  \\
\gls{Time-series} step changes & & & & & & & & & & & & & & & \tick & & &  \\
\gls{Time-series} predictability measure & & & & & & & & & & & & & & & \tick & & & \\
\gls{Time-series} non-linearity measure & & & & & & & & & & & & & & & \tick & & & \\
\gls{Time-series} largest Lyapunov exponent & & & & & & & & & & & & & & \tick & \tick & & & \\
\gls{Time-series} 3 largest power spectrum frequencies & & & & & & & & & & & & & & & \tick & & & \\
\gls{Time-series} maximum value of power spectrum & & & & & & & & & & & & & & & \tick & & & \tick  \\
\gls{Time-series} number of peaks $>$ 60\% & & & & & & & & & & & & & & & \tick & & & \\
\gls{Time-series} auto-correlations at lags 1 and 2 & & & & & & & & & & & & & & & \tick & & &  \\
\gls{Time-series} partial auto-correlations at lags 1 and 2 & & & & & & & & & & & & & & & \tick & & &  \\
\gls{Time-series} seasonality Measure & & & & & & & & & & & & & &\tick & \tick & & &  \\
\gls{Time-series} mean \gls{SMAPE} - mean deviated \gls{SMAPE} & & & & & & & & & & & & & & & \tick & & & \\
\gls{Time-series} mean \gls{SMAPE} / mean deviated \gls{SMAPE} & & & & & & & & & & & & & & & \tick & & &  \\
\gls{Time-series} mean of correlation coefficient & & & & & & & & & & & & & & & \tick & & &  \\
\gls{Time-series} \gls{StdDev} of correlation coefficient & & & & & & & & & & & & & & & \tick & & & \\
\gls{Time-series} methods in top performing cluster & & & & & & & & & & & & & & & \tick & & & \\
\gls{Time-series} distance top performing cluster to second best & & & & & & & & & & & & & & & \tick & & & \\
\gls{Time-series} Serial \gls{CORR} Box-Pierce statistic & & & & & & & & & & & & & & \tick\footnote{of raw and trend/seasonally adjusted} & & & & \\
\gls{Time-series} Non-linear autoregressive structure & & & & & & & & & & & & & & \tick\footnote{of raw and trend/seasonally adjusted} & & & & \\
\gls{Time-series} Self-similarity (Long-range Dependence & & & & & & & & & & & & & & \tick & & & & \\
\gls{Time-series} Periodicity (frequency) & & & & & & & & & & & & & & \tick & & & & \\
Min. of \gls{CORR} between predictors and target & & & & & & & & & & & & & \tick & & & & & \\
Max. of \gls{CORR} between predictors and target & & & & & & & & & & & & & \tick & & & & & \\
Mean of \gls{CORR} between predictors and target & & & & & & & & & & & & & \tick & & & & & \\
\gls{StdDev} of absolute value of \gls{CORR} between predictors and target & & & & & & & & & & & & & \tick & & & & &  \\
Min. of \gls{CORR} between pairs of predictors & & & & & & & & & & & & & \tick & & & & & \\
Max. of \gls{CORR} between pairs of predictors & & & & & & & & & & & & & \tick & & & & & \\
Mean of \gls{CORR} between pairs of predictor & & & & & & & & & & & & & \tick & & & & & \\
\gls{StdDev} of absolute value of \gls{CORR} between pairs of predictors & & & & & & & & & & & & & \tick & & & & & \\ \hline

\multicolumn{19}{l}{Information Theoretic Meta-features}  \\ \hline
\gls{HC} & & & \tick & \tick & & \tick & & & & & \tick & & & & & & & \\
Entropy of nominal features & & & \tick & \tick & & \tick & & & & & & & & & & & & \\
\gls{HCX} & & & \tick & \tick & & \tick & & & & & & & & & & & & \\
\gls{MCX} & & & \tick & \tick & & \tick & & & & & \tick & & & & & & & \\
Class Entropy to Mutual information ratio & & & & \tick & & \tick & & & & & & & & & & & & \tick \\
\gls{NoiseRaio} & \tick & & & \tick & & \tick & & & & & & & & & & & &  \\
Dispersion Gain & & & & & & & & & & & & & & & & & \tick &  \\ \hline

\multicolumn{19}{l}{Landmarkers} \\ \hline
\gls{DecisionNodes} & & & & & & \tick & & \tick & & & & & & & & \tick & & \\
\gls{WorstNodes} & & & & & & \tick & & & & & & & & & & \tick & & \\
\gls{RandomlyNodes} & & & & & & \tick & & & & & & & & & & \tick & & \tick \\
\gls{NB} & & & & & & \tick & & \tick & & & & & & & & \tick & & \tick \\
\gls{k-NN} & & & & \tick\footnote{k = 3 used only in \cite{Giraud-Carrier2005}} & & \tick\footnote{k = 1}\addtocounter{footnote}{-1}\addtocounter{Hfootnote}{-1} & \tick & \tick\footnotemark\addtocounter{footnote}{-1}\addtocounter{Hfootnote}{-1} & & \tick & & & & & & \tick\footnotemark & & \tick  \\
\gls{e-NN} & & & & & & \tick & & & & & & & & & & & &  \\
\gls{LDA} & & & & & & \tick & & \tick & & & & & & & & \tick & & \tick \\
\gls{C5.0trees} & & & & & & & & \tick & & & & & & & & & & \tick \\
\gls{C5.0boost} & & & & & & & \tick & \tick & & & & & & & & & & \\
\gls{C5.0rules} & & & & & & & \tick & \tick & & \tick & & & & & & & & \\
\gls{Ripper} & & & & & & & \tick & & & & & & & & & & & \\
\gls{Ltree} & & & & & & & \tick & & & \tick & & & & & & & & \\
\gls{AverageNodes} & & & & & & & & & & & & & & & & \tick & & \\ \hline

\multicolumn{19}{l}{Model-based Meta-features}  \\ \hline
Nodes per attribute & & & & & \tick & & & & & & & & & & & & & \\
Nodes per instance & & & & & \tick & & & & & & & & & & & & & \\
Average leaf corroboration & & & & & \tick & & & & & & & & & & & & & \\
Average gain-ratio difference & & & & & \tick & & & & & & & & & & & & & \\
Maximum depth & & & & & \tick & & & & & & & & & & & & & \\
No. of repeated nodes & & & & & \tick & & & & & & & & & & & & & \\
Shape & & & & & \tick & & & & & & & & & & & & & \\
Homogeneity & & & & & \tick & & & & & & & & & & & & & \\
Imbalance & & & & & \tick & & & & & & & & & & & & &  \\
Internal symmetry & & & & & \tick & & & & & & & & & & & & &  \\
No. of Nodes in each level - width & & & & & & & & & \tick & & & & & & & & &  \\
No. of levels - Height & & & & & & & & & \tick & & & & & & & & &  \\
No. of nodes in the tree & & & & & & & & & \tick & & & & & & & & &  \\
No. of leaves in the tree & & & & & & & & & \tick & & & & & & & & &  \\
Maximum no. of nodes at one level & & & & & & & & & \tick & & & & & & & & &  \\
Mean of the no. of nodes & & & & & & & & & \tick & & & & & & & & &  \\
\gls{StdDev} of the no. of nodes & & & & & & & & & \tick & & & & & & & & &  \\
Length of the Shortest branch & & & & & & & & & \tick & & & & & & & & &  \\
Length of the Longest branch & & & & & & & & & \tick & & & & & & & & &  \\
Mean of the branch length & & & & & & & & & \tick & & & & & & & & &  \\
\gls{StdDev} of the branch length & & & & & & & & & \tick & & & & & & & & &  \\
Minimum occurrence of Features & & & & & & & & & \tick & & & & & & & & &  \\
Maximum occurrence of Features & & & & & & & & & \tick & & & & & & & & & \\
Mean of the no. of occurrences of Features & & & & & & & & & \tick & & & & & & & & & \\
\gls{StdDev} of no. of occurrences of Features & & & & & & & & & \tick & & & & & & & & & \\

Weight sum of dataset & & & & & & & & & & & & & & & & & & \\
Minimum weight sum of dataset & & & & & & & & & & & & & & & & & & \\
Average weight sum of dataset & & & & & & & & & & & & & & & & & & \\
\gls{StdDev} weight sum of dataset & & & & & & & & & & & & & & & & & & \\

No. neighbours for dataset & & & & & & & & & & & & & & & & & & \\
Minimum No. neighbours for dataset & & & & & & & & & & & & & & & & & & \\
Maximum No. neighbours for dataset & & & & & & & & & & & & & & & & & & \\
Average No. neighbours for dataset & & & & & & & & & & & & & & & & & & \\
\gls{StdDev} of No. neighbours for dataset & & & & & & & & & & & & & & & & & & \\

\gls{PCA} 95\% & & & & & & & & & & & & & & & & & & \tick \\
\gls{PCA} skewness & & & & & & & & & & & & & & & & & & \tick \\
\gls{PCA} kurtosis & & & & & & & & & & & & & & & & & & \tick \\ \hline

Total Meta-features & 9 & 13 & 19 & 25 & 10 & 14 & 8 & 7 & 15 & 3 & 7 & 11 & 10 & 9 & 23 & 7 & 10 & 22 \\
\bottomrule
\end{longtable}}
\end{center}

\newpage
\bibliographystyle{unsrt}  
\bibliography{references}  

\end{document}